\pdfoutput=1

\documentclass[11pt]{article}

\usepackage[]{EMNLP2022}
\usepackage{times}

\usepackage[T1]{fontenc}

\usepackage{amsmath,amsfonts,bm}

\def\eqref#1{equation~\ref{#1}}

\def\1{\bm{1}}

\DeclareMathAlphabet{\mathsfit}{\encodingdefault}{\sfdefault}{m}{sl}
\SetMathAlphabet{\mathsfit}{bold}{\encodingdefault}{\sfdefault}{bx}{n}

\def\sW{{\mathbb{W}}}

\usepackage[utf8]{inputenc}

\usepackage{microtype}

\usepackage{inconsolata}
\usepackage{times}
\usepackage{latexsym}
\usepackage[T1]{fontenc}

\usepackage[utf8]{inputenc}

\usepackage{microtype}
\usepackage{comment}
\usepackage{todonotes}
\usepackage{color,soul}
\usepackage{graphicx}
\usepackage{pbox}
\usepackage{subcaption}
\usepackage{epstopdf}
\usepackage{xstring}
\usepackage{fdsymbol}
\usepackage{multirow}
\usepackage{mathtools}
\usepackage[normalem]{ulem}
\usepackage{pifont}
\usepackage{todonotes}
\usepackage{tikz-dependency}
\usepackage{diagbox}
\usepackage{array}
\usepackage{lipsum}
\usepackage[normalem]{ulem}
\usepackage{booktabs}
\useunder{\uline}{\ul}{}
\usepackage{enumitem,kantlipsum}
\usepackage{mathtools}

\newcommand{\zz}{\mathbf{z}}

\newcommand{\modelM}{\mathbb{M}}

\usepackage{comment}

\title{On the Transformation of Latent Space in Fine-Tuned NLP Models \\\hphantom{...}
\\\footnotesize{\textcolor{red}{WARNING: This paper contains model outputs which may be disturbing to the reader}}}

\author{
 Nadir Durrani$^{\diamondsuit}$ \hspace{2mm} Hassan Sajjad$^{\clubsuit}$\thanks{\hspace{1.5mm} This work was carried out while the author was at QCRI.} \hspace{2mm} Fahim Dalvi$^{\diamondsuit}$ \hspace{2mm} Firoj Alam$^{\diamondsuit}$    \\
 $^{\diamondsuit}$Qatar Computing Research Institute, Hamad Bin Khalifa University, Qatar \\
 $^\clubsuit$Faculty of Computer Science, Dalhousie University, Canada \\
 {\{ndurrani,faimaduddin, fialam\}@hbku.edu.qa}, \hspace{1mm} hsajjad@dal.ca
}

\begin{document}
\maketitle
\begin{abstract}

We study the evolution of latent space in fine-tuned NLP models. Different from the commonly used probing-framework, we opt for an unsupervised method to analyze representations. More specifically, we discover latent concepts in the representational space using hierarchical clustering. We 
then use an alignment function to gauge the similarity between the latent space of a pre-trained model and its fine-tuned version. We use traditional linguistic concepts to facilitate our understanding and also study how the model space transforms towards task-specific information. We perform a thorough analysis, comparing pre-trained and fine-tuned models across three models and three downstream tasks. The notable findings of our work are: i) the latent space of the higher layers evolve towards task-specific concepts, ii) whereas the lower layers retain generic concepts acquired in the pre-trained model, iii) we discovered that some concepts in the higher layers acquire polarity towards the output class, and iv) that these concepts can be used for generating adversarial triggers.


\end{abstract}

\section{Introduction}



The revolution of deep learning models in NLP can be attributed to 
transfer learning from pre-trained language models.
Contextualized representations learned within these 
models capture rich linguistic knowledge that can be leveraged towards novel tasks 
e.g. classification of COVID-19 tweets \cite{Alam_covid_infodemic_2021,valdes-etal-2021-uach}, disease prediction \cite{medBERT} 
or natural language understanding tasks such as SQUAD \cite{rajpurkar-etal-2016-squad} and GLUE \cite{wang-etal-2018-glue}.  

Despite their success, the 
opaqueness of deep neural networks remain a cause of concern and has spurred a new area of research to analyze these models. A large body of work 
analyzed the knowledge learned within representations of pre-trained models~\cite{belinkov:2017:acl,conneau2018you,liu-etal-2019-linguistic,tenney-etal-2019-bert, durrani-etal-2019-one,rogers-etal-2020-primer} and 
showed the presence of core-linguistic knowledge in various parts of the network. 
Although transfer learning using pre-trained models has become ubiquitous, very few papers~\cite{merchant-etal-2020-happens, mosbach-etal-2020-interplay, durrani-etal-2021-transfer} have 
analyzed the representations of the fine-tuned models. 
Given their massive usability, interpreting fine-tuned models and highlighting task-specific peculiarities is critical for their deployment in real-word scenarios, where it is important to ensure fairness and trust 
when applying AI  solutions.

In this paper, we focus on analyzing fine-tuned models and investigate:
\textit{how does the latent space evolve in a fine-tuned model?} Different from the commonly used probing-framework of training a post-hoc classifier \cite{belinkov:2017:acl,dalvi:2019:AAAI}, we opt for an unsupervised method to analyze the latent space of pre-trained models. More specifically, we cluster contextualized representations in high dimensional space using hierarchical clustering  
and term these clusters as the \emph{Encoded Concepts} \cite{dalvi2022discovering}. 
We then analyze how these encoded concepts evolve 
as the models are fine-tuned towards a downstream task. Specifically, we target the following questions: \emph{i) how do the latent spaces compare between base\footnote{We use ``base'' and ``pre-trained'' models interchangeably.} and the fine-tuned models?} \emph{ii) how does the presence of core-linguistic concepts change during transfer learning?} and \emph{iii) how is the knowledge of downstream tasks 
structured in a fine-tuned model?} 


We use an alignment function \cite{sajjad:naacl:2022} to compare the 
concepts encoded in the fine-tuned models with: i) the 
concepts encoded in their pre-trained base models, ii) the human-defined concepts (e.g. parts-of-speech tags or semantic properties), and iii) the labels of the downstream task towards which the model is fine-tuned.

\begin{figure*}[!ht]
    \centering
    \includegraphics[width=1.0\linewidth]{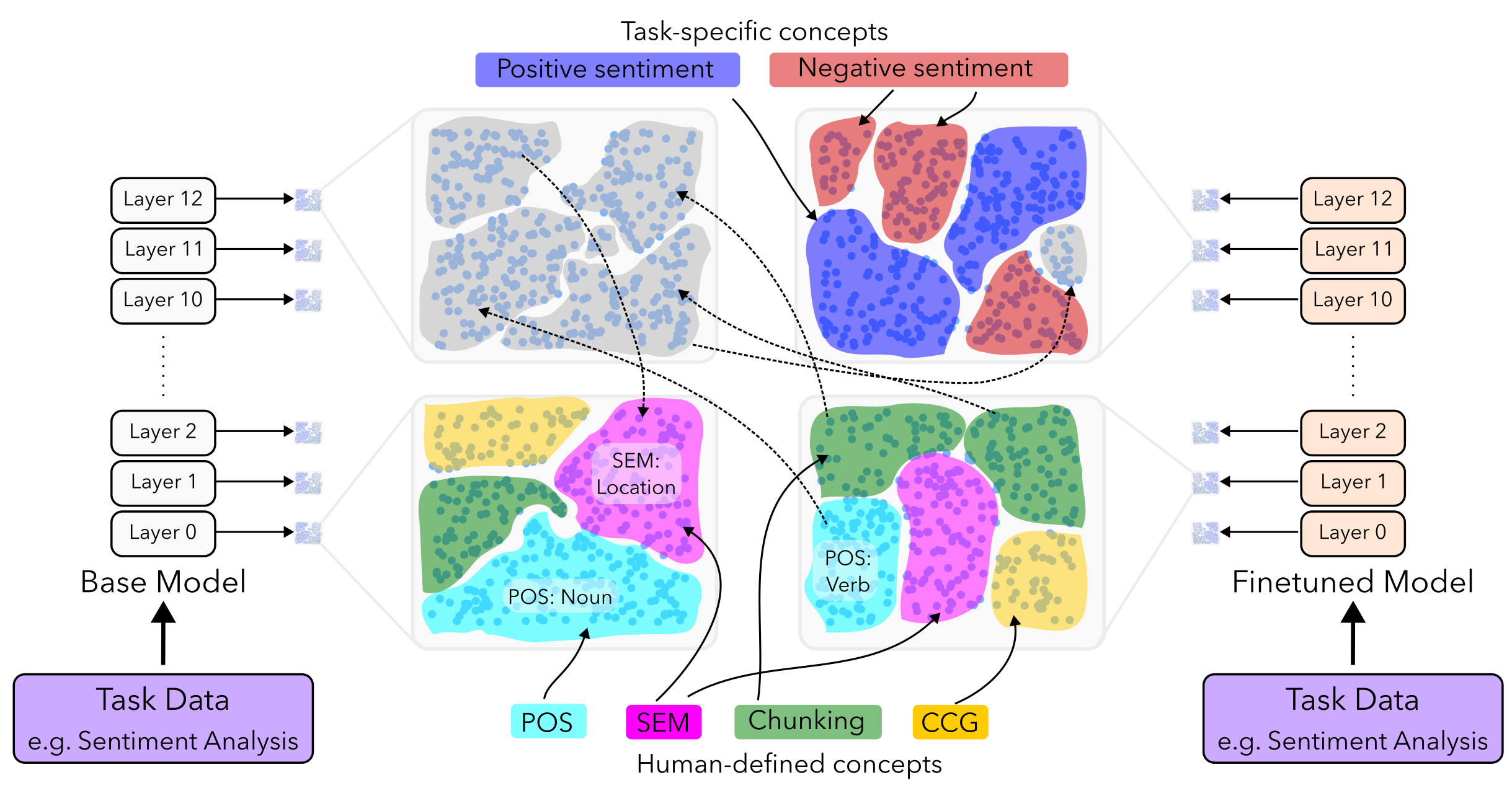}
    \caption{Comparing encoded concepts of a model across different layers with: i) the  concepts encoded its base model (dashed lines), ii) human-defined concepts (e.g. POS tags or semantic properties), and iii) task specific concepts (e.g. positive or negative sentiment class).
    }
    \label{fig:pipeline}
\end{figure*}
We carried out our study using three pre-trained transformer language models; BERT~\cite{devlin-etal-2019-bert}, XLM-RoBERTa~\cite{xlm-roberta} and ALBERT~\cite{lan2019albert}, analyzing how their representation space evolves as they are fine-tuned towards the task of Sentiment Analysis~\cite[SST-2,][]{socher-etal-2013-recursive}, Natural Language Inference~\cite[MNLI,][]{williams-etal-2018-broad} and Hate Speech Detection~\cite[HSD,][]{hateXplain}. Our 
analysis yields interesting insights such as:

\begin{itemize}
    \item The latent space of the models 
    substantially evolve 
    from their base versions after fine-tuning. 
    \item The latent space representing core-linguistic concepts is limited to the lower layers in the fine-tuned models, contrary to the base models where it is distributed across the network.
    \item 
    We found task-specific polarity concepts in the higher layers of the Sentiment Analysis and Hate Speech Detection tasks.
    \item These polarized concepts can be 
    used as triggers to generate adversarial examples. 
    \item Compared to BERT and XLM, the representational space in ALBERT changes significantly 
    during fine-tuning. 
\end{itemize}

\section{Methodology}
\label{sec:methodology}

Our work builds on 
the Latent Concept Analysis method \cite{dalvi2022discovering} for interpreting 
representational spaces of neural network models. We cluster contextualized embeddings to discover \emph{Encoded Concepts} in the model and study the evolution of the latent space in the fine-tuned model by aligning the encoded concepts of the fine-tuned model to: i) their pre-trained version, ii) 
the human-defined concepts and iii) 
the task-specific concepts 
(for the task the pre-trained model is fine-tuned on). Figure \ref{fig:pipeline} presents an overview of our approach. In the following, we define the scope of \textit{Concept} and discuss each step of our approach in detail.



\subsection{Concept}
\label{sec:concept}

We 
define concept as a group of words that are clustered together based on any linguistic relation such
as lexical, semantic, syntactic, morphological etc.
Formally, consider $C_t(n)$ as a concept consisting of a unique set of words $\{w_1, w_2, \dots, w_J\}$ where $J$ is the number of words in $C_t$, $n$ is a concept identifier, and $t$ is the concept type which can be an encoded concept ($ec$), a human-defined concept ($pos:verbs, sem:loc, \dots$) and a class-based concept ($sst:+ive, hsd:toxic, \dots$).

\begin{figure*}[t]
    \begin{subfigure}[b]{0.31\linewidth}
    \centering
    \includegraphics[width=\linewidth]{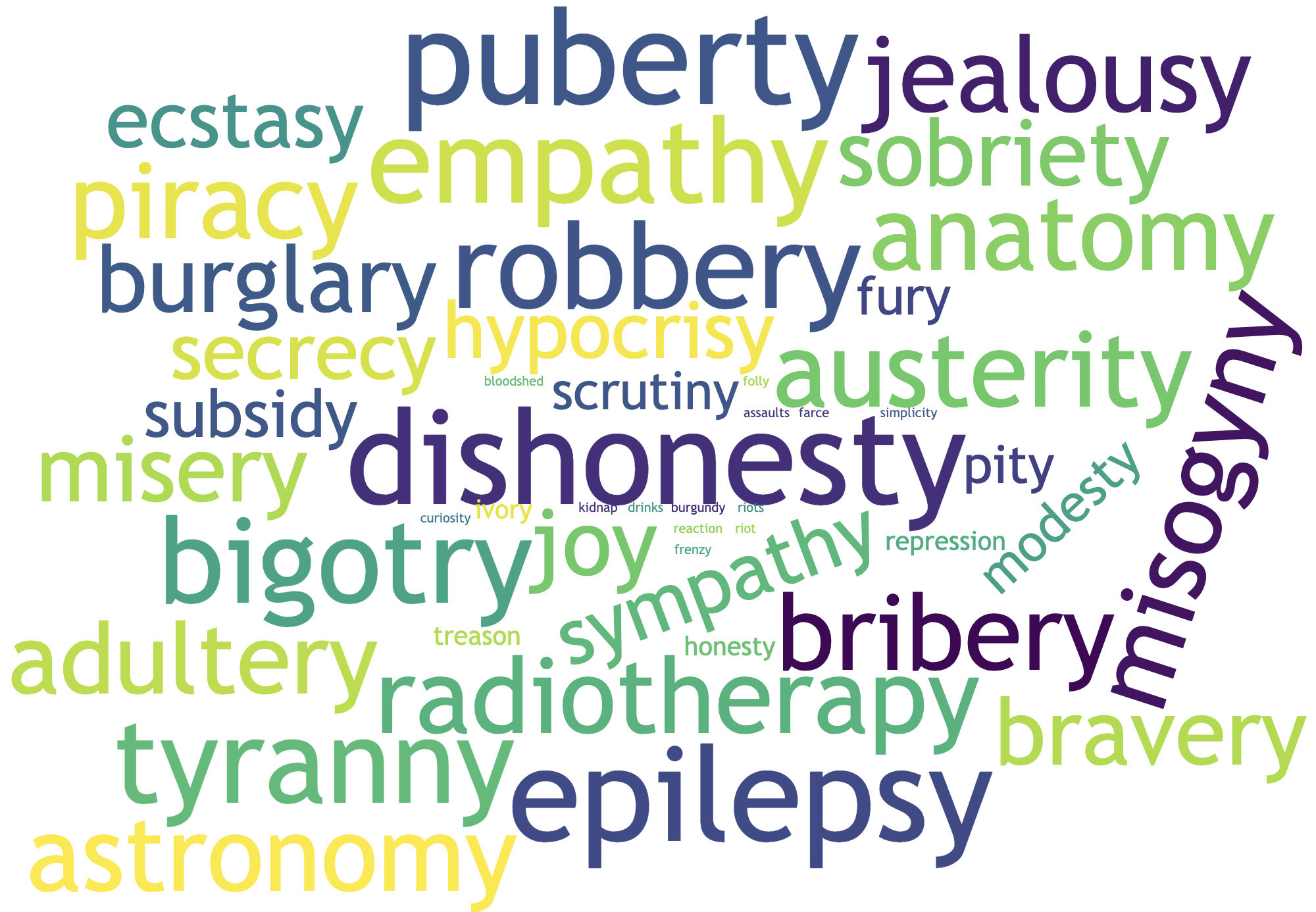}
    \caption{Nouns ending with ``y''}
    \label{fig:nouns+y}
    \end{subfigure}
    \begin{subfigure}[b]{0.31\linewidth}
    \centering
    \includegraphics[width=\linewidth]{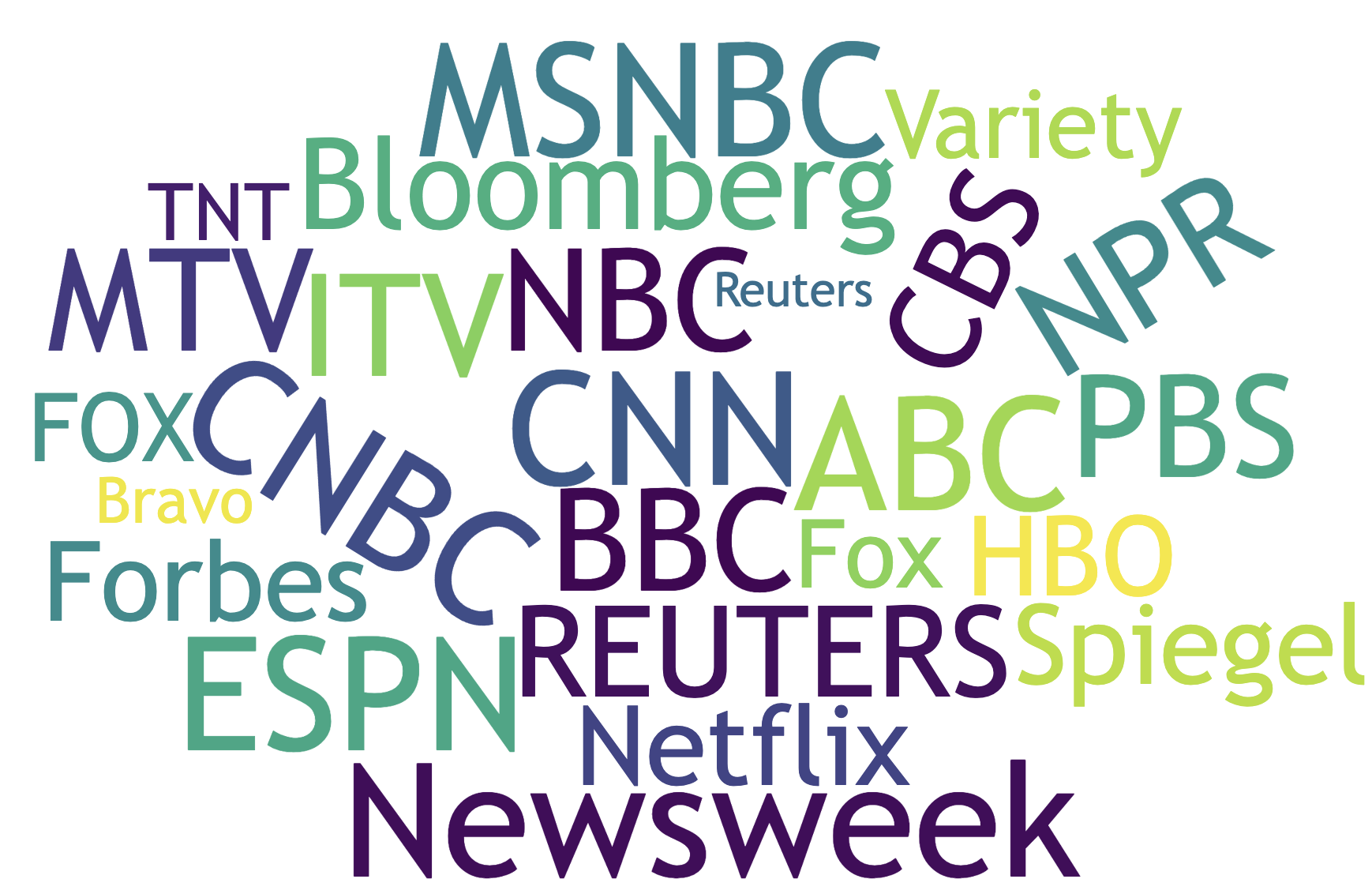}
    \caption{Named Entities -- TV}
    \label{fig:tv}
    \end{subfigure}
    \begin{subfigure}[b]{0.31\linewidth}
    \centering
    \includegraphics[width=\linewidth]{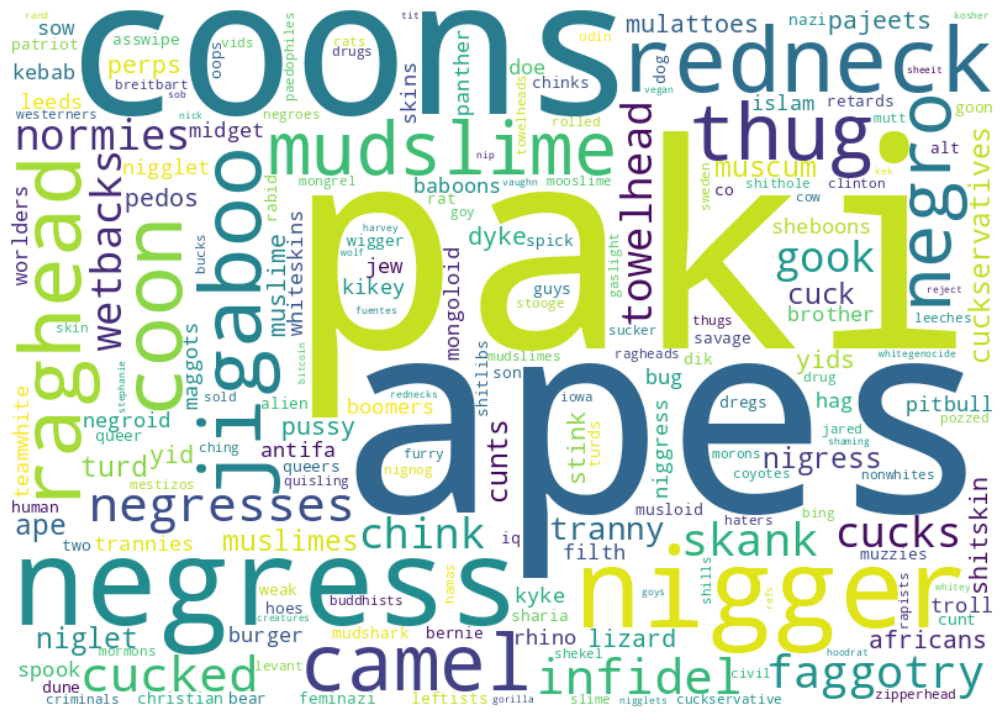}
    \caption{Racial Slurs}
    \label{fig:racial-slurs}
    \end{subfigure}

    \caption{Examples of encoded concepts. The size of a specific word is based on its frequency in the cluster, defined by the number of times different contextual representations of a word were grouped in the same cluster.}
    \label{fig:sample_clusters}
\end{figure*}

\paragraph{Encoded Concepts:}

Figure~\ref{fig:sample_clusters} 
shows a few examples of the encoded concepts discovered in the BERT model, where the concept is defined by 
a group based on nouns ending with ``y'' (Figures~\ref{fig:nouns+y}) or 
a group based on TV related named entities (Figure~\ref{fig:tv}).  
Similarly, Figure \ref{fig:racial-slurs} 
is a concept representing racial slurs in a BERT model tuned for Hate Speech Detection (HSD) task.
We 
denote this concept 
as $C_{ec}(\texttt{bert-hsd-layer10-c227}) = \{paki, nigger, mudslime, redneck \dots\}$, 
i.e. the concept was discovered in the layer 10 of the BERT-HSD model and $c227$ is the concept number.

\paragraph{Human Concepts:} Each individual tag in the human-defined concepts such as parts-of-speech (POS), semantic tagging (SEM) represents a concept $C$. For example, $C_{pos} (JJR) = \{greener, taller, happier, \dots \}$ defines a concept containing comparative adjectives in the POS tagging task, $C_{sem} (MOY) = \{January, February, \dots, December\}$ defines a concept containing months of the year in the semantic tagging task.

\paragraph{Task-specific Concepts:}

Another 
kind of concept that we use in this 
work is the task-specific concepts where the concept represents affinity of its members with respect to the task labels. Consider a sentiment classification task with two labels ``positive'' and ``negative''. We define $C_{sst} (+ve)$ as a concept containing words 
when they only appear in sentences that are labeled positive. 
Similarly, we define $C_{hsd} (toxic)$ as a concept that contain words that only appear in the sentences that were marked as toxic.


\subsection{Latent Concept Discovery}
\label{sec:clustering}

A vector representation in the neural network
model is composed of feature attributes of the input words. We group the encoded vector representations using a clustering approach discussed below. The underlying clusters, that we term as
the encoded concepts, are then matched with the
human-defined concepts using an alignment function.

Formally, consider a pre-trained model $\mathbf{M}$ with $L$ layers: $\{l_1, l_2, \ldots, l_L\}$. Given a dataset $\sW=\{w_1, w_2, ..., w_N\}$, we generate feature vectors, a sequence of latent representations: $\sW\xrightarrow{\modelM}\zz^l = \{\zz^l_1, \dots, \zz^l_n\}$\footnote{Each element $z_i$ denotes contextualized word representation for the corresponding word $w_i$ in the sentence.} by doing a forward-pass on the data for any given layer $l$. 
Our goal is to cluster representations $\zz^l$, from task-specific training data to obtain \emph{encoded concepts}.

We use agglomerative hierarchical clustering~\citep{gowda1978agglomerative}, which assigns each word to its individual cluster and iteratively combines the clusters based on Ward's minimum variance criterion, using intra-cluster variance. Distance between two vector representations is calculated with the squared Euclidean distance. The algorithm terminates when the required $K$ clusters (i.e. encoded concepts) are formed, where $K$ is a hyper-parameter. Each encoded concept represents a latent relationship between the words present in the cluster. 

\subsection{Alignment}
\label{sec:alignment}

Once we have obtained a set of encoded concepts in the base (pre-trained) and fine-tuned models, we want to align them to study how the latent space has evolved during transfer learning. \newcite{sajjad:naacl:2022} calibrated representational space in transformer models with different linguistic concepts to generate their explanations. We extend their alignment function to align latent spaces within a model and its fine-tuned version. Given a concept $C_1(n)$ with $J$ number of words, we consider it to be $\theta$-aligned ($\Lambda_{\theta}$) with 
concept $C_2(m)$, if they satisfy the following constraint:

%
%
%

\begin{equation}
  \Lambda_{\theta}(C_1, C_2)=\left\{
  \begin{array}{@{}ll@{}}
    1, & \text{if}\ \frac{\sum_{w\in C_1} \sum_{w' \in C_2} \delta(w,w')}{J} \geq \theta  \\
    0, & \text{otherwise},
  \end{array}\right.
  \label{eq:al}
\end{equation} 

\noindent where Kronecker function $\delta(w,w')$ is defined as 

\begin{equation*}
  \delta(w,w')=\left\{
  \begin{array}{@{}ll@{}}
    1, & \text{if}\ w=w' \\
    0, & \text{otherwise}
  \end{array}\right.
\end{equation*} 


\paragraph{Human-defined Concepts} The function can be used to draw a mapping between concepts different types of discussed in Section \ref{sec:concept}. To investigate \emph{how the transfer learning impacts human-defined knowledge}, we align the latent space to the human-defined concepts such as $C_{pos}(NN)$ or $C_{chunking} (PP)$. 

\paragraph{Task Concepts} Lastly, we compare the encoded concepts with the task-specific concepts. 
Here, we use the alignment function to mark affinity of an encoded concept. 
For the Sentiment Analysis task, let a task-specific concept $C_{sst}(+ve) = \{w^+_1, w^+_2,\dots,w^+_n\}$ defined by a set words that only appeared in positively labeled sentences $S=\{s^+_1, s^+_2, \dots, s^+_n\}$. We call a concept $C_{ec} = \{x_1, x_2, \dots, x_n\}$ aligned to $C_{sst} (+ve)$ and mark it positive if 
all words ($\geq \theta$) in the encoded concept appeared in positively labeled sentences. Note that here a word represents an instance based on its contextualized embedding. We similarly align $C_{ec}$ with $C_{sst}(-ve)$ to discover negative polarity concepts.

\section{Experimental Setup}
\label{sec:setup}

\subsection{Models and Tasks}
\label{sec:tasks}

We experimented with three popular transformer architectures namely: BERT-base-cased \cite{devlin-etal-2019-bert}, XLM-RoBERTa \cite{xlm-roberta} and ALBERT (v2) \cite{lan2019albert} using the base versions (13 layers and 768 dimensions). 
To carryout the analysis, we fine-tuned the base models for the tasks of sentiment analysis using the Stanford sentiment treebank dataset \cite[SST-2,][]{socher-etal-2013-recursive}, natural language inference \cite[MNLI,][]{williams-etal-2018-broad} and the Hate Speech Detection task~\cite[HSD,][]{hateXplain}.

\subsection{Clustering}
\label{sec:clusteringSettings}
We used the task-specific training data for clustering using both the base (pre-trained) and fine-tuned models. This enables to accurately compare the representational space generated by the same data. We do a forward-pass over both base and fine-tuned models to generate contextualized feature vectors\footnote{We use NeuroX toolkit \cite{dalvi2019neurox} to extract contextualized representations.} of words in the data and run agglomerative hierarchical clustering over these vectors. We do this for every layer independently, obtaining $K$ clusters (a.k.a encoded concepts) for both base and fine-tuned models. We used $K=600$ for our experiments.\footnote{We experimented with ELbow~\cite{Thorndike53whobelongs} and  Silhouette~\cite{silhouetteCluster} methods to find the optimal number of clusters, but could not observe a reliable pattern. Selecting between $600-1000$ clusters gives the right balance to avoid over-clustering (many small clusters) and under-clustering (a few large clusters).} We carried out preliminary experiments (all the BERT-base-cased experiments) using $K=200, 400, \dots, 1000$  and all our experiments using $K=600$ and $K=1000$. We found that our results are not sensitive to these parameters and the patterns are consistent with different cluster settings (please see Appendix~\ref{sec:appendix:clustering}).

\begin{figure*}[t]
    \centering
    \includegraphics[width=\linewidth]{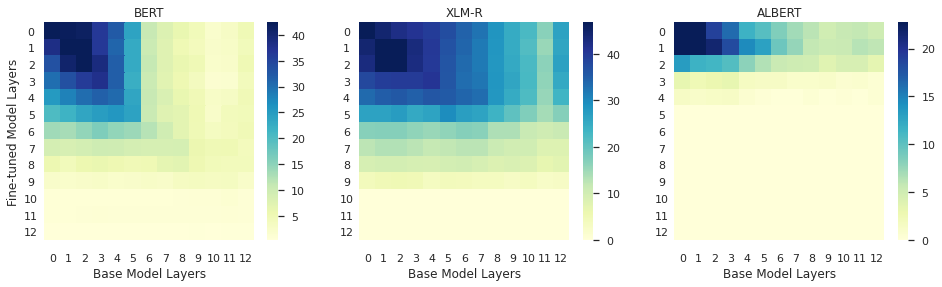}
    \caption{Comparing encoded concepts 
    of base models with their SST fine-tuned versions. X-axis = base model, Y-axis = fine-tuned model. Each cell in the matrix represents a percentage (aligned concepts/total concepts in a layer) between the base and fine-tuned models. Darker color means higher percentage. Detailed plots with actual overlap values are provided in the Appendix.}
    \label{fig:basevsft}
\end{figure*}

\subsection{Human-defined Concepts}
\label{sec:humanConcepts}

We experimented with traditional tasks that are defined to capture core-linguistic concepts such as word morphology: part-of-speech tagging using the Penn TreeBank data \cite{marcus-etal-1993-building}, syntax: chunking tagging  using CoNLL 2000 shared task dataset \cite{tjong-kim-sang-buchholz-2000-introduction}, CCG super tagging using CCG Tree-bank \cite{hockenmaier2006creating} and semantic tagging using the Parallel Meaning Bank data \cite{abzianidze-EtAl:2017:EACLshort}. We trained BERT-based sequence taggers for each of the above tasks and annotate the task-specific training data. Each core-linguistic task serves as a human-defined concept that is aligned with encoded concepts to measure the representation of linguistic knowledge in the latent space. 
Appendix \ref{sec:appendix:linguisticConcepts} presents the details on human defined concepts, data stats and tagger accuracy.

\subsection{Alignment Threshold}
\label{sec:alignmentThreshold}
We consider an encoded concept to be aligned with another concept, if it has at least 95\%\footnote{Using an overlap of $\geq95\%$ provides a very tight threshold, allowing only $5\%$ of noise. Our patterns were consistent at lower and higher thresholds.}
 match in the number of words. We only consider concepts that have more than 5 word-types. Note that the encoded concepts are based on contextualized embedding where a word has different embeddings depending on the context. 

\section{Analysis}
\label{sec:analysis}


Language model pre-training has been shown to capture rich linguistic features \cite{tenney-etal-2019-bert, belinkov-etal-2020-analysis} that are redundantly distributed across the network \cite{dalvi-2020-CCFS,durrani-etal-2020-analyzing}. We analyze how the representational space transforms when tuning towards a downstream task: i) how much knowledge is carried forward and ii) how it is redistributed, using our alignment framework.


\subsection{Comparing Base and Fine-tuned Models}
\label{subsec:base}

\emph{How do the latent spaces compare between base and fine-tuned models?}
We measure the 
overlap between the concepts encoded in the different layers of the base and fine-tuned models to guage the extent of transformation.  Figures~\ref{fig:basevsft} compares 
the concepts in the base BERT, XLM-RoBERTa and ALBERT 
models versus their fine-tuned variants on the SST-2 task.\footnote{Please see all results in Appendix \ref{sec:appendix:base}.} We observe a high overlap in concepts in the lower layers of the model that starts decreasing as we go deeper in the network, completely diminishing towards the end. We conjecture that
\textbf{\emph{the lower layers of the model retain generic language concepts learned in the base model,
where as the higher layers are now learning task-specific concepts.}}\footnote{Our next results comparing the latent space with human-defined language concepts (Section \ref{subsec:linguistic}) and the task specific concepts (Section \ref{subsec:taskspecific}) reinforces this hypothesis.}
Note, however, that the lower layers also do not completely align between the models, which shows that all the layers go through substantial changes during transfer learning. 

\paragraph{Comparing Architectures:} The spread of the shaded area along the x-axis, particularly in XLM-R, reflects that some higher layer latent concepts 
in the base model have shifted towards the lower layers of the fine-tuned model. 
The latent space in the higher layers 
now reflect task-specific knowledge 
which was not present in the base model. 
ALBERT 
shows a strikingly different pattern with only the first 2-3 layers exhibiting an overlap with base concepts. This could be attributed to the fact that ALBERT shares parameters across layers while the other models have separate parameters for every layer. 
ALBERT has less of a luxury to preserve previous knowledge and therefore its space transforms significantly towards the downstream 
task. Notice that the overlap is 
comparatively smaller (38\% vs. 52\% and 46\% compared to BERT and XLM-R respectively), even in the embedding layer, where the words are primarily grouped based on lexical similarity. 

\begin{figure*}[t]
    \begin{subfigure}[b]{0.33\linewidth}
    \centering
    \includegraphics[width=\linewidth]{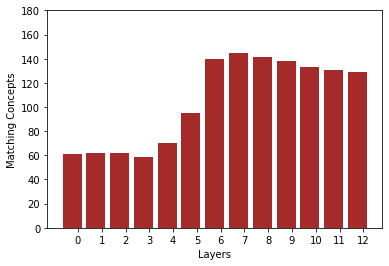}
    \caption{BERT -- Base}
    \label{fig:bert}
    \end{subfigure}
    \begin{subfigure}[b]{0.33\linewidth}
    \centering
    \includegraphics[width=\linewidth]{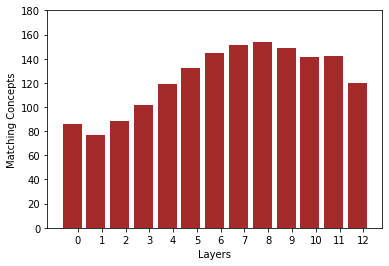}
    \caption{XLM-R -- Base}
    \label{fig:xlm-r}
    \end{subfigure}
    \begin{subfigure}[b]{0.33\linewidth}
    \centering
    \includegraphics[width=\linewidth]{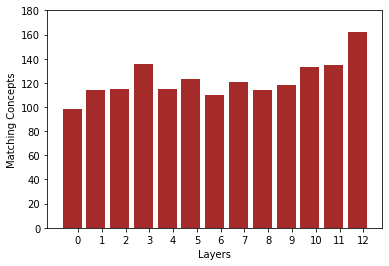}
    \caption{ALBERT -- Base}
    \label{fig:albert}
    \end{subfigure}
    \begin{subfigure}[b]{0.33\linewidth}
    \centering
    \includegraphics[width=\linewidth]{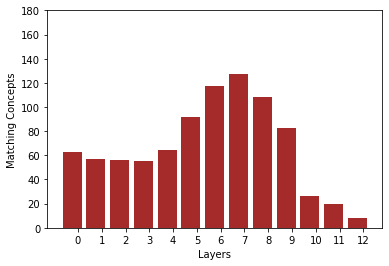}
    \caption{BERT -- SST}
    \label{fig:bert}
    \end{subfigure}
    \begin{subfigure}[b]{0.33\linewidth}
    \centering
    \includegraphics[width=\linewidth]{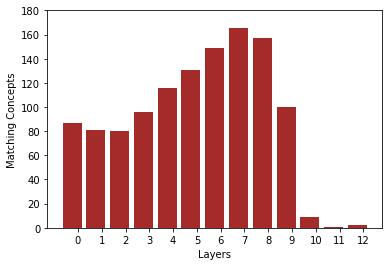}
    \caption{XLM-R -- SST}
    \label{fig:xlm-r}
    \end{subfigure}
    \begin{subfigure}[b]{0.33\linewidth}
    \centering
    \includegraphics[width=\linewidth]{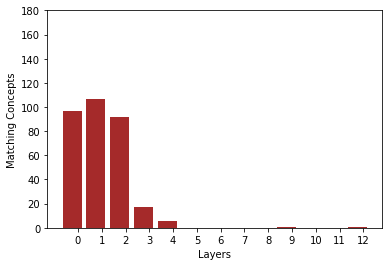}
    \caption{ALBERT -- SST}
    \label{fig:albert}
    \end{subfigure}
    \caption{Alignment of the encoded concepts with POS concepts (e.g., determiners, past-tense verbs, superlative adjectives) in the base and fine-tuned SST models. The maximum possible concepts per layer are 600 (total \# of clusters). Note that the POS information depreciates significantly in the final layers in the SST-tuned models.}
    \label{fig:humanConcepts}
\end{figure*}

\begin{figure}[t]
    \begin{subfigure}[b]{0.48\linewidth}
    \centering
    \includegraphics[width=\linewidth]{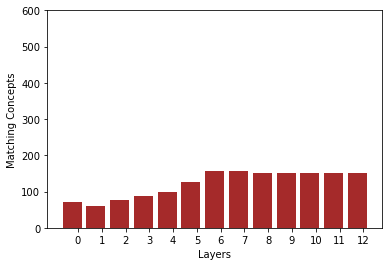}
    \caption{BERT -- Base}
    \label{fig:bert-base}
    \end{subfigure}
    \begin{subfigure}[b]{0.48\linewidth}
    \centering
    \includegraphics[width=\linewidth]{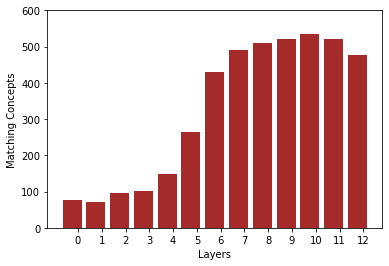}
    \caption{BERT -- POS}
    \label{fig:bert-pos}
    \end{subfigure}
    \caption{Alignment of the the encoded concepts with POS in the BERT base versus fine-tuned POS models. In contrast to the results in Figure \ref{fig:humanConcepts}, POS concepts appreciate significantly when the model is tuned towards the POS task. At most 23\% concepts align in the BERT-base model as opposed to BERT-pos where close to 84\% encoded concepts are aligned to the POS tags.}
    \label{fig:humanConcepts-pp}
\end{figure}

\subsection{Presence of  Linguistic Concepts in the Latent Space}
\label{subsec:linguistic}

\textit{How does the presence of core-linguistic concepts change during transfer learning?} To validate our hypothesis that generic language concepts are now predominantly retained in the lower half, we analyze 
how the linguistic concepts spread across the layers in the pre-trained and fine-tuned models by aligning the latent space to the human-defined concepts.
Figure~\ref{fig:humanConcepts} shows that the latent space of the models capture POS concepts (e.g., determiners, past-tense verbs, superlative adjectives etc.) The information is 
present across the layers in the pre-trained models, however, as the model is fine-tuned towards downstream task, it is retained only at the lower and middle layers. 
We can draw two conclusions from this result: i) POS information is important for a foundational task such as language modeling (predicting the masked word), but not critically important for a sentence classification task like sentiment analysis. To strengthen our argument and confirm this further, we fine-tuned a BERT model towards the task of POS tagging itself. Figure \ref{fig:humanConcepts-pp} shows the extent of the alignment between POS concept with BERT-base and BERT tuned models towards the POS. 
Notice that more than 80\% encoded concepts in the final layers of the BERT-POS 
model are now aligned with the POS concept as opposed to the BERT-SST model where POS concept (as can be seen in Figure~\ref{fig:humanConcepts}) decreased to less than 5\%. 


\paragraph{Comparing Tasks and Architectures}
We found these observations to be consistently true for other tasks (e.g., MNLI and HSD) and human-defined concepts (e.g., SEM, Chunking and CCG tags) across the three architectures (i.e., BERT, XLM-R and ALBERT) that we study in this paper.\footnote{Please see Appendix \ref{subsec:appendix:linguistic} for all the results.} Table~\ref{tab:conceptDrop} compares an overall presence of core-linguistic concepts across the base and fine-tuned models. We observe a consistent deteriorating pattern across all human-defined concepts. In terms of architectural difference we again found ALBERT to show a substantial difference in the representation of POS post fine-tuning. The number of concepts not only regressed to the lower layers, but also decreased significantly as opposed to the base model.  



\begin{table}[]
\centering
\footnotesize
\begin{tabular}{l| rrrr }
\toprule
\textbf{Tasks} & \textbf{POS} & \textbf{SEM} & \textbf{Chunking} & \textbf{CCG} \\ \midrule

%
BERT(B)       & 17.6 & 23.5 & 27.3 &  20.6\\
BERT(SST) & 11.2 & 17.0 & 21.4 &  15.1 \\
\midrule
XLM-R(B)        & 20.6 & 22.6 & 22.1 &  17.9 \\
XLM-R(SST) & 15.1 & 18.0 & 19.9 &  15.2 \\
\midrule
ALBERT(B)        & 20.4 & 27.3 & 32.6 &  25.6 \\
ALBERT(SST) & 4.1 & 5.3 & 8.2 & 4.1 \\
\bottomrule

\end{tabular}
\caption{Overall presence (percentage of aligned concepts) of human-defined concepts in base (B) versus SST fine-tuned models.}
\label{tab:conceptDrop}
\end{table}

\begin{figure*}[t]
    \begin{subfigure}[b]{0.33\linewidth}
    \centering
    \includegraphics[width=\linewidth]{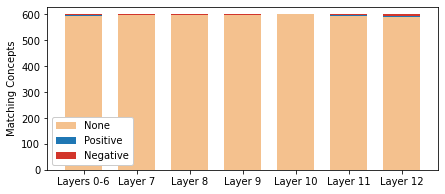}
    \caption{BERT -- Base}
    \label{fig:bert-base}
    \end{subfigure}
    \begin{subfigure}[b]{0.33\linewidth}
    \centering
    \includegraphics[width=\linewidth]{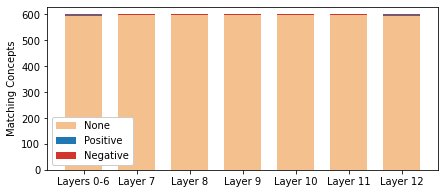}
    \caption{XLM-R -- Base}
    \label{fig:xlm-r-base}
    \end{subfigure}
    \begin{subfigure}[b]{0.33\linewidth}
    \centering
    \includegraphics[width=\linewidth]{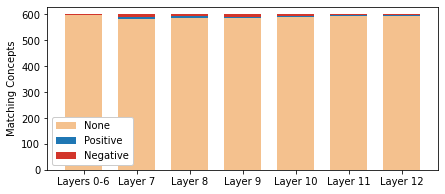}
    \caption{ALBERT -- Base}
    \label{fig:albert-base}
    \end{subfigure}
    \begin{subfigure}[b]{0.33\linewidth}
    \centering
    \includegraphics[width=\linewidth]{Figures/BERT-SST-SST}
    \caption{BERT -- SST}
    \label{fig:bert-sst}
    \end{subfigure}
    \begin{subfigure}[b]{0.33\linewidth}
    \centering
    \includegraphics[width=\linewidth]{Figures/XLM-SST-SST}
    \caption{XLM-R -- SST}
    \label{fig:xlm-r-sst}
    \end{subfigure}
    \begin{subfigure}[b]{0.33\linewidth}
    \centering
    \includegraphics[width=\linewidth]{Figures/ALBERT-SST-SST}
    \caption{ALBERT -- SST}
    \label{fig:albert-sst}
    \end{subfigure}
    \caption{Aligning encoded concepts with the task specific concepts in Base and their corresponding SST tuned models.}
    \label{fig:conceptPolarity}
\end{figure*}

\begin{figure}[t]
    
    \begin{subfigure}[b]{0.48\linewidth}
    \centering
    \includegraphics[width=\linewidth]{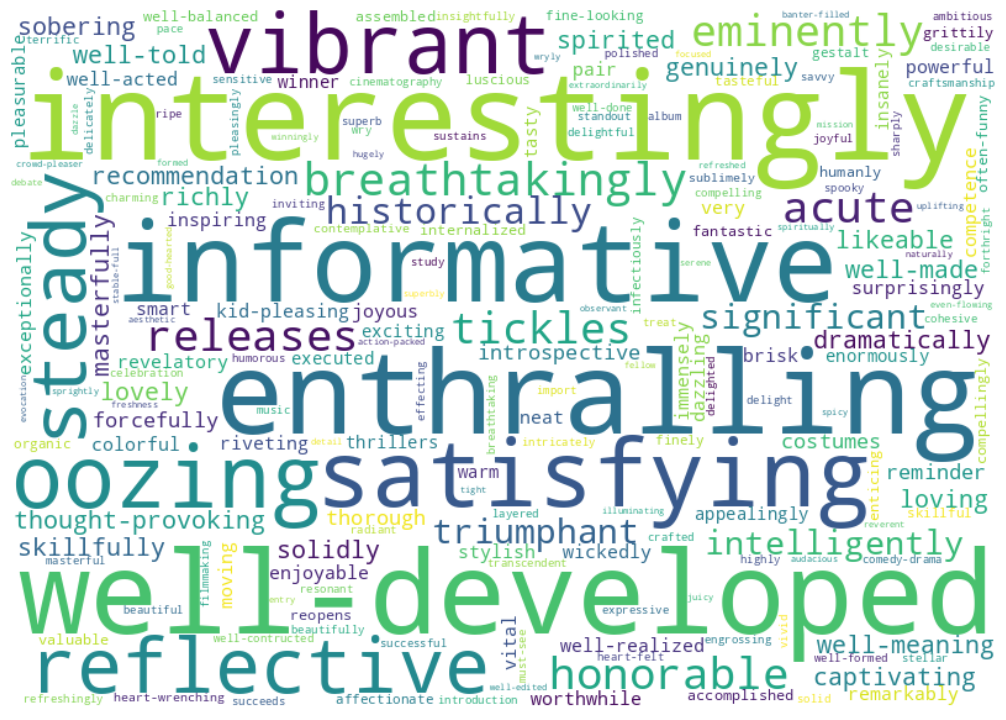}
    \label{fig:xlm-positive}
    \vspace{-4mm}
    \caption{XLM-SST L12, c15}
    \end{subfigure}
    \begin{subfigure}[b]{0.48\linewidth}
    \centering
    \includegraphics[width=\linewidth]{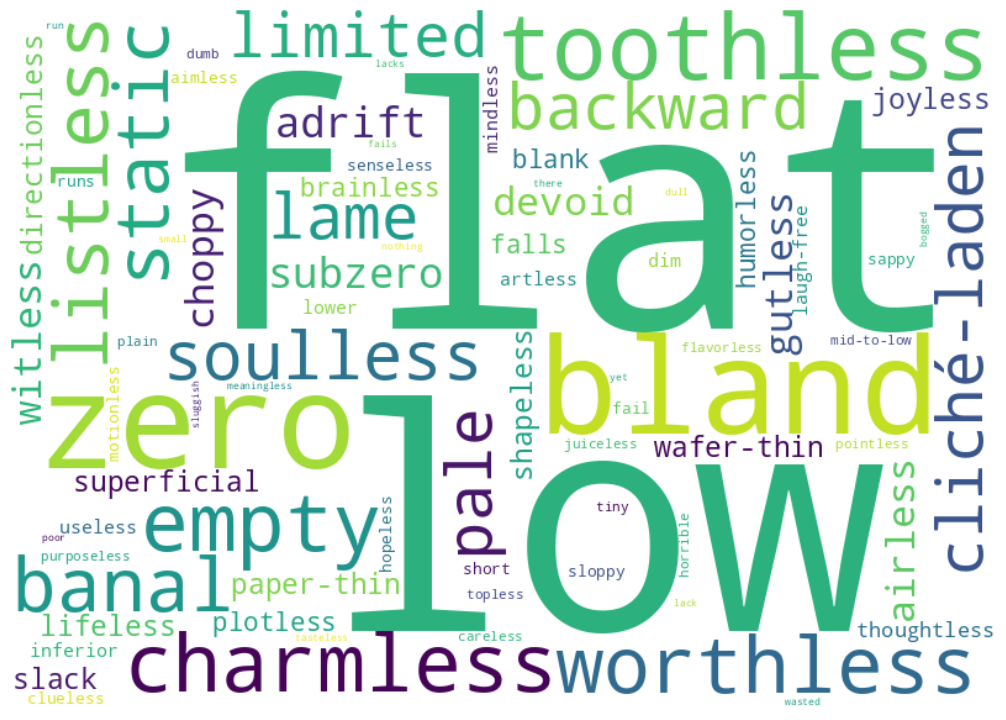}
    \label{fig:xlm-negative}
    \vspace{-4mm}
    \caption{XLM-SST L10, c16}
    \end{subfigure}
    \begin{subfigure}[b]{\linewidth}
    \centering
    \vspace{2mm}
    \includegraphics[width=\linewidth]{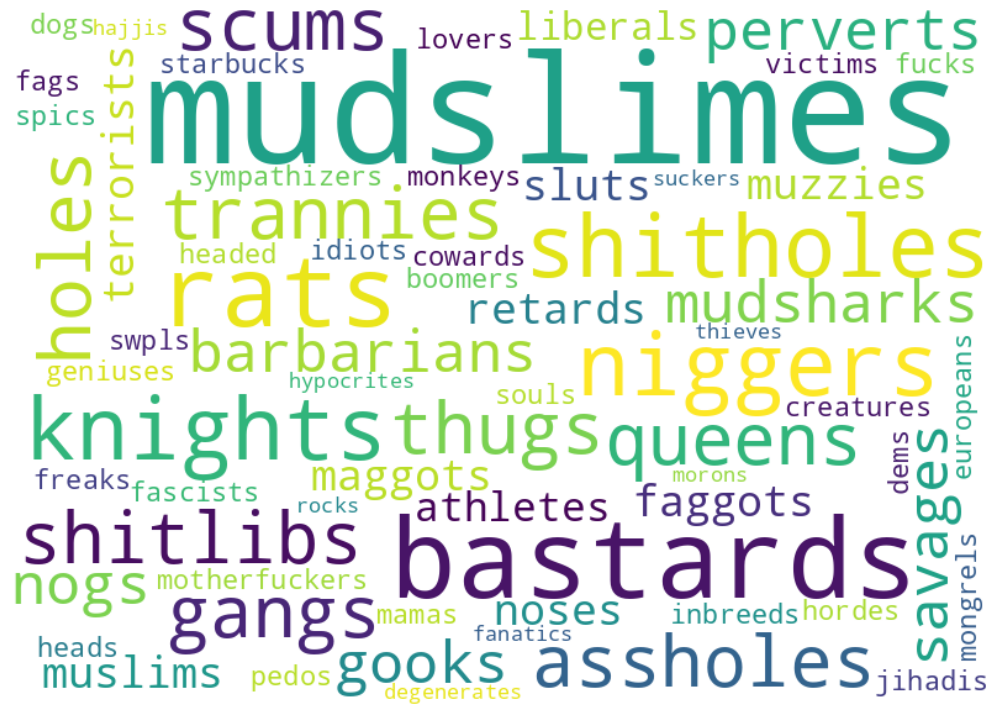}
    \label{fig:xlm-toxic}
    \vspace{-4mm}
    \caption{XLM-HSD L10, c576}
    \end{subfigure}
    \vspace{-4mm}
    \caption{Polarity Concepts in XLM-R models: Positive (top left) and Negative (top right) in the SST task, Toxic Concept (bottom) in the HSD task.}
    \label{fig:sample_polarity_clusters}
    \vspace{-2mm}
\end{figure}

\subsection{Task-specific Latent Spaces}
\label{subsec:taskspecific}

\textit{How 
is the knowledge of downstream tasks 
structured in a fine-tuned models?}  Now that we have established  
that the latent space of higher layers are substantially different from base models and from linguistic concepts, we probe: \textit{what kind of knowledge is learned in the latent space of higher layers?} 
Previous research ~\cite{kovaleva-etal-2019-revealing,merchant-etal-2020-happens, durrani-etal-2021-transfer} found that
the higher layers are optimized for the task. 
We also noticed how the concepts learned in the top 6 layers of the BERT-POS 
model completely evolve towards the (POS) task labels (See Figure \ref{fig:humanConcepts-pp}). We now extend this experiment towards the sentence-level tasks and 
investigate the extent of alignment between latent concepts of the fine-tuned models with its task labels. 
The SST-2 task 
predicts the sentiment (positive or negative) of a sentence. Using the class label, we form positive and negative polarity concepts, 
and align the polarity concepts with the encoded concepts.\footnote{Positive polarity concept is made up of words that only appeared in the positively labeled sentences. We say an encoded concept ($C_{ec}$) is aligned to positive polarity concept ($C^+$) if $\geq95\%$ words in $C_{ec} \in C^+$. Note that the opposite is not necessarily true.} If an encoded concept is not aligned with any polarity concept, we 
mark the concept as ``Neutral''. 
Figure~\ref{fig:conceptPolarity} shows that the concepts in the final layers acquire polarity towards the task of output classes compared to the base model 
where we only see neutral concepts throughout the network.
Figure \ref{fig:sample_polarity_clusters} shows an example of positive (top left) and negative polarity (top right) concepts in the XLM-R model tuned for the SST task. 
The bottom half shows a toxic concept in the model trained towards the HSD task. Please see Appendix \ref{sec:appendix:sample_clusters} for more examples.

\paragraph{Comparing architectures} Interestingly, the presence of polarity clusters is not always equal. The last two layers of BERT-SST are dominated by negative polarity clusters, 
while ALBERT showed an opposite trend where the positive polarity concepts were more frequent. 
We 
hypothesized that the imbalance in the presence of polarity clusters may reflect 
prediction bias towards/against a certain class. However, we did not find a clear evidence for this in a pilot experiment. We collected predictions for all three models over a random corpus of 37K sentences. The models predicted negative sentiment by 69.5\% (BERT), 67.4\% (XLM) and 64.4\% (ALBERT). While the numbers weakly correlate with the number of negative polarity concepts in these models, a thorough investigation is required to obtain accurate insights. We leave a detailed exploration of this for future.

ALBERT showed the evolution of polarity clusters much earlier in the network (Layer 3 onwards). This is inline with our previous results on aligning encoded concepts of base and fine-tuned models (Figure~\ref{fig:basevsft}). We found that the latent space in ALBERT evolved the most, overlapping with its base model only in the first 2-3 layers. The POS-based concepts were also reduced just to the first two layers (Figure~\ref{fig:humanConcepts}). Here we can see that the concepts learned within the remaining layers acquire affinity towards the task specific labels. We found these results to be consistent with the hate speech task (See Appendix \ref{sec:appendix:taskLabels}) but not in the MNLI task, where we did not find the latent concepts to acquire affinity towards the task labels. This could be attributed to the complexity and nature of the MNLI task.
Unlike the SST-2 and HSD tasks, where lexical triggers serve as an important indicators for the model, MNLI requires intricate modeling of semantic relationships between premise and hypothesis to predict entailment. Perhaps an alignment function that models the interaction between 
the concepts of premise and hypothesis is required. We leave this exploration for the future.




\section{Adversarial Triggers}
\label{sec:triggers}

The discovery of polarized concepts in the SST-2 and HSD tasks, motivated us to question:  
\textbf{\emph{whether the fine-tuned model is learning the actual task or relying on lexical triggers to solve the problem.}} Adversarial examples have been used in the literature to highlight model's vulnerability \cite{kuleshov2018adversarial, wallace-etal-2019-universal}. We show that our polarity concepts can be used to generate such examples using the following formulation:

Let $C_{ec}(+ve) = \{C^{+}_{1}, C^{+}_{2}, \dots, C^{+}_{N}\}$ be a set of latent concepts that are identified to have a strong affinity towards predicting positive sentiment in the SST task. Let $S^{-} = \{s^{-}_{1}, s^{-}_{2}, \dots, s^{-}_{M} \}$ be the sentences in a dev-set that are predicted as negative by the model. We compute the flipping accuracy of each concept $C^{+}_{x}$ using the following function:
	
\vspace{-2mm}
$$F(C^{+}_{x}, S^{-}) = \frac{1}{N_a} \sum_{w_i \in C^{+}_{x}} \sum_{s_{j} \in S_{-}} \gamma(w_i,s_j)$$

\noindent where $\gamma(w_i,s_j) = 1$, if prepending $w_i$ 
to the sentence $s_j$ flips the model's prediction from negative to positive. Here $N_a$ is the total number of adversarial examples that were generated, and equates to $|C^{+}_{x}|\times|S^{-}|$. We similarly compute the flipping accuracy $F(C^{-}_{x}, S^{+})$ of the concepts that acquire affinity towards the negative class. The concepts with high flipping accuracy can be used to generate adversarial examples.


\begin{table}[]
\centering
\footnotesize
\begin{tabular}{r| rrr }
\toprule
\textbf{Tasks} & \textbf{Layer 10} & \textbf{Layer 11} & \textbf{Layer 12} \\ 
\midrule
& \multicolumn{3}{c}{\textbf{BERT SST }} \\
$+$ve $\rightarrow$ $-$ve & 43.6 & 41.2 & 43.4 \\
$-$ve $\rightarrow$ $+$ve & 41.0 & 42.1 & 44.7 \\
\midrule
& \multicolumn{3}{c}{\textbf{XLM-RoBERTa SST}} \\
$+$ve $\rightarrow$ $-$ve & 42.8 & 41.7 & 43.0  \\
$-$ve $\rightarrow$ $+$ve & 29.7 & 31.1 & 33.7 \\
\midrule
& \multicolumn{3}{c}{\textbf{ALBERT SST}} \\
$+$ve $\rightarrow$ $-$ve & 69.2 & 73.8 & 77.2  \\
$-$ve $\rightarrow$ $+$ve & 69.6 & 74.2 & 70.9 \\
\midrule
& \multicolumn{3}{c}{\textbf{BERT HS }} \\
nt $\rightarrow$ tx & 65.2 & 41.5 & 59.3 \\
tx $\rightarrow$ nt & 11.2 & 6.74 & 8.91  \\
\midrule
& \multicolumn{3}{c}{\textbf{XLM-RoBERTa HS}} \\
nt $\rightarrow$ tx & 57.7 & 69.0 & 38.9 \\
tx $\rightarrow$ nt & 7.23 & 9.14 & 9.60  \\
\midrule
& \multicolumn{3}{c}{\textbf{ALBERT HS}} \\
nt $\rightarrow$ tx & 84.9 & 65.4 & 91.5\\
tx $\rightarrow$ nt & 0.00 & 0.00 & 0.00  \\
\bottomrule

\end{tabular}
\caption{Flipping accuracy (\%age) of top-5 polarized concepts: $+$ve $\rightarrow$ $-$ve = flipping a positive sentence to negative using negative polarity concept, nt $\rightarrow$ tx = converting a non-toxic sentence toxic using toxic concept.}
\label{tab:adversarial}
\vspace{-2mm}
\end{table}

We compute the flipping accuracy of each polarized concept on a small hold-out set. Table \ref{tab:adversarial} shows the average flipping accuracy of the top-5 polarized concepts for each class (positive/negative in SST-2 and toxic/non-toxic in the HSD task) across final three layers on the test-set. We observed that by just prepending the words in highly polarized concepts, we are able to effectively flip the model's prediction by up to $91.5\%$. This shows that \textbf{\emph{these models are fragile and heavily rely on lexical triggers to make predictions}}. In the case of Hate Speech Detection task, we observed that while it is easy to make a non-toxic sentence toxic, it is hard to reverse the affect. 

\paragraph{Comparing Architectures} We found ALBERT to be an outlier once again with a high flipping accuracy, which shows that ALBERT relies on these cues more than the other models and is therefore more prone to adversarial attacks. 

\section{Related Work}
\label{sec:relatedWork}

A plethora of papers have
been written in the past five years on interpreting deep NLP models. The work done in this direction can be broadly classified into: i) post-hoc representation analysis that encode the contextualized embedding for the knowledge learned \cite{dalvi:2017:ijcnlp, belinkov-etal-2020-analysis,rogers-etal-2020-primer,lepori-mccoy-2020-picking} and ii) causation analysis that connect input features with model behavior as a whole and at a level of individual predictions \cite{linzen_tacl, gulordava-etal-2018-colorless, marvin-linzen-2018-targeted}.\footnote{Please read \cite{belinkov-glass-2019-analysis,neuronSurvey} for comprehensive surveys of methods.} Our work mainly falls in the former category although we demonstrated a causal link between the encoded knowledge and model predictions by analyzing the concepts in the final layers and demonstrating how they can be used to generate adversarial examples with lexical triggers. Recent work \cite{feder-etal-2021-causalm, elazar-etal-2021-amnesic} attempts to bridge the gap by connecting the two lines of work.



Relatively less work has been done on interpreting fine-tuned models. \newcite{zhao-bethard-2020-berts} analyzed the heads encoding negation scope in fine-tuned BERT and RoBERTa models. \newcite{merchant-etal-2020-happens,mosbach-etal-2020-interplay} analyzed linguistic knowledge in pre-trained models and showed that while fine-tuning changes the upper layers of the model, but does not lead to ``catastrophic forgetting of linguistic phenomena''. Our results resonate with their findings, in that the higher layers learn task-specific concepts.\footnote{Other works such as \cite{ethayarajh-2019-contextual,sajjad2023:csl} have also shown higher layers to capture task-specific information.} However, similar to \newcite{durrani-etal-2021-transfer} we found depreciation of linguistic knowledge in the final layers. \newcite{mehrafarin-etal-2022-importance} showed that the size of the datasets used for fine-tuning should be taken into account to draw reliable conclusions when using probing classifiers. A pitfall to the probing classifiers is the difficulty to disentangle probe's capacity to learn from the actual knowledge learned within the representations \cite{hewitt-liang-2019-designing}.  
Our work is different from all the previous work done on interpreting fine-tuned models. We do away from the limitations of probing classifiers by using an unsupervised approach.

Our work is inspired by the recent work on discovering latent spaces for analyzing pre-trained models \cite{michael-etal-2020-asking, dalvi2022discovering, Yao-Latent, sajjad:naacl:2022}.
Like \newcite{dalvi2022discovering, sajjad:naacl:2022} 
we discover encoded concepts in pre-trained models and align them with pre-defined concepts. Different from them, we study the evolution of latent spaces of fine-tuned models. 

\section{Conclusion}
\label{sec:conclusion}
We studied the evolution of latent space of pre-trained models when fine-tuned towards a downstream task. Our approach uses hierarchical clustering to find encoded concepts in the representations. We 
analyzed them by comparing with the encoded concepts of base model,  human-defined concepts, and task-specific concepts. We showed that the latent space of fine-tuned model is substantially different from their base counterparts. The human-defined linguistic knowledge largely vanishes from the higher layers. The higher layers encode task-specific concepts relevant to solve the task. Moreover, we showed that these task-specific concepts can be used in generating adversarial examples that flips the predictions of the model up to 91\% of the time in the case of ALBERT Hate Speech model. The discovery of word-level task-specific concepts suggest that the models rely on lexical triggers and are vulnerable to adversarial attacks.


\section{Limitations}
\label{sec:limitations}

The hierarchical clustering is memory intensive. For instance, the clustering of 250k representation vectors, each of size 768 consumes 400GB of CPU memory. This limits the applicability of our approach to small to medium data sizes. Moreover, our approach is limited to word-level concepts. The models may also learn phrasal concepts to solve a task. We speculate that the low matches of affinity concepts in the MNLI task is due to the limitation of our approach in analyzing phrasal units.



\bibliography{anthology,custom}
\bibliographystyle{acl_natbib}

\newpage
\clearpage
\section*{Appendix}
\label{sec:appendix}
\appendix

\begin{figure*}[h]
    \begin{subfigure}[b]{0.33\linewidth}
    \centering
    \includegraphics[width=\linewidth]{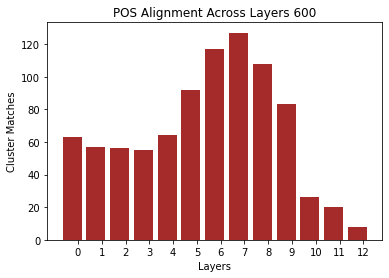}
    \caption{BERT -- SST (600)}
    \label{fig:bert}
    \end{subfigure}
    \begin{subfigure}[b]{0.33\linewidth}
    \centering
    \includegraphics[width=\linewidth]{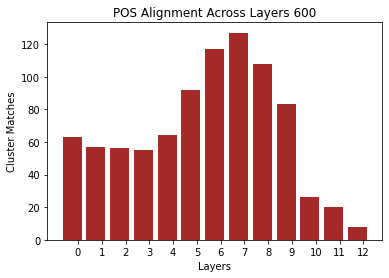}
    \caption{XLM-R -- SST (600)}
    \label{fig:xlm-r}
    \end{subfigure}
    \begin{subfigure}[b]{0.33\linewidth}
    \centering
    \includegraphics[width=\linewidth]{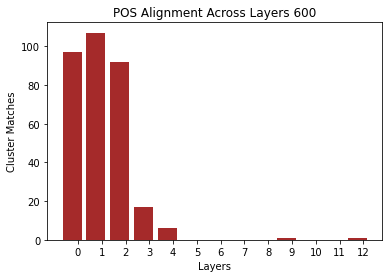}
    \caption{ALBERT -- SST (600)}
    \label{fig:albert}
    \end{subfigure}
    \begin{subfigure}[b]{0.33\linewidth}
    \centering
    \includegraphics[width=\linewidth]{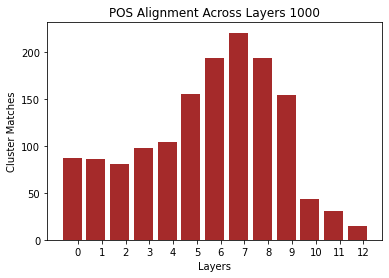}
    \caption{BERT -- SST (1000)}
    \label{fig:bert}
    \end{subfigure}
    \begin{subfigure}[b]{0.33\linewidth}
    \centering
    \includegraphics[width=\linewidth]{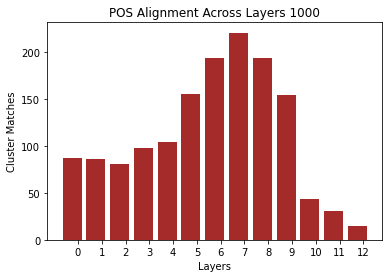}
    \caption{XLM-R -- SST (1000)}
    \label{fig:xlm-r}
    \end{subfigure}
    \begin{subfigure}[b]{0.33\linewidth}
    \centering
    \includegraphics[width=\linewidth]{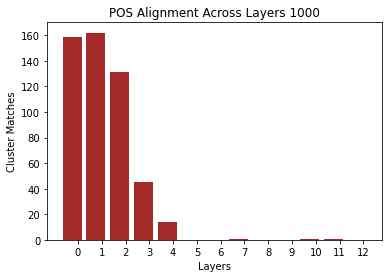}
    \caption{ALBERT -- SST (1000)}
    \label{fig:albert}
    \end{subfigure}
    \caption{Comparing encoded concepts when using 600 or 1000 clusters}
    \label{fig:appendix:humanConcepts-pos}
\end{figure*}

\begin{figure*}[h]
    \begin{subfigure}[b]{0.31\linewidth}
    \centering
    \includegraphics[width=\linewidth]{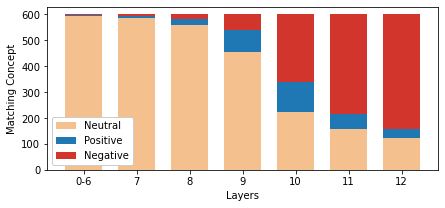}
    \caption{BERT -- SST (600)}
    \label{fig:bert}
    \end{subfigure}
    \begin{subfigure}[b]{0.31\linewidth}
    \centering
    \includegraphics[width=\linewidth]{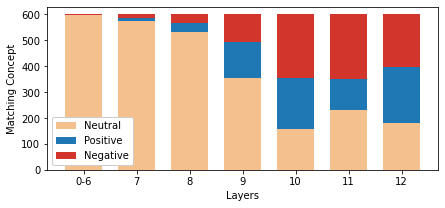}
    \caption{XLM-R -- SST (600)}
    \label{fig:xlm-r}
    \end{subfigure}
    \begin{subfigure}[b]{0.31\linewidth}
    \centering
    \includegraphics[width=\linewidth]{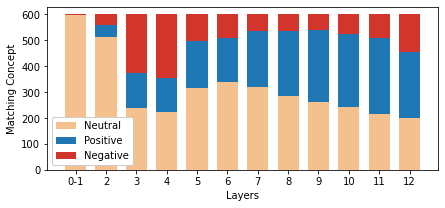}
    \caption{ALBERT -- SST (600)}
    \label{fig:albert}
    \end{subfigure}
    \begin{subfigure}[b]{0.33\linewidth}
    \centering
    \includegraphics[width=\linewidth]{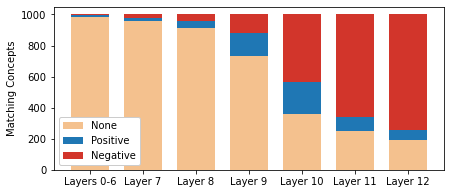}
    \caption{BERT -- SST (1000)}
    \label{fig:bert}
    \end{subfigure}
    \begin{subfigure}[b]{0.33\linewidth}
    \centering
    \includegraphics[width=\linewidth]{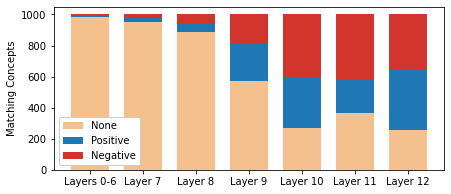}
    \caption{XLM-R -- SST (1000)}
    \label{fig:xlm-r}
    \end{subfigure}
    \begin{subfigure}[b]{0.33\linewidth}
    \centering
    \includegraphics[width=\linewidth]{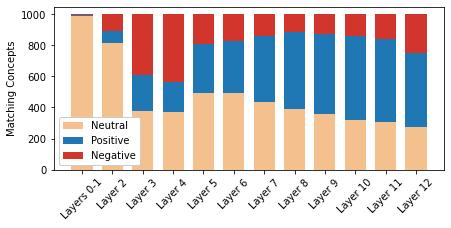}
    \caption{ALBERT -- SST (1000)}
    \label{fig:albert}
    \end{subfigure}
    \caption{Aligning encoded concepts with the task specific concepts}
    \label{fig:appendix:conceptPolarity-1000:sst}
\end{figure*}

\section{Linguistic Concepts}
\label{sec:appendix:linguisticConcepts}

We used parts-of-speech tags (48 concepts) using Penn Treebank data \cite{marcus-etal-1993-building}, semantic tags (73 concepts) \cite{abzianidze-EtAl:2017:EACLshort}, chunking tags \cite{tjong-kim-sang-buchholz-2000-introduction} (22 concepts) and CCG super tags (1272 concepts). Please see all the concepts below. This provides a good coverage of linguistic concepts including morphology, syntax and semantics.

\begin{table}[!tbh]
\centering
\scalebox{0.75}{
\setlength{\tabcolsep}{2.0pt}
\begin{tabular}{@{}lll@{}}
\toprule
\textbf{\#} & \textbf{Tag} & \textbf{Description} \\ \midrule
1 & CC & Coordinating conjunction \\
2 & CD & Cardinal number \\
3 & DT & Determiner \\
4 & EX & Existential there \\
5 & FW & Foreign word \\
6 & IN & Preposition or subordinating conjunction \\
7 & JJ & Adjective \\
8 & JJR & Adjective, comparative \\
9 & JJS & Adjective, superlative \\
10 & LS & List item marker \\
11 & MD & Modal \\
12 & NN & Noun, singular or mass \\
13 & NNS & Noun, plural \\
14 & NNP & Proper noun, singular \\
15 & NNPS & Proper noun, plural \\
16 & PDT & Predeterminer \\
17 & POS & Possessive ending \\
18 & PRP & Personal pronoun \\
19 & PRP\$ & Possessive pronoun \\
20 & RB & Adverb \\
21 & RBR & Adverb, comparative \\
22 & RBS & Adverb, superlative \\
23 & RP & Particle \\
24 & SYM & Symbol \\
25 & TO & to \\
26 & UH & Interjection \\
27 & VB & Verb, base form \\
28 & VBD & Verb, past tense \\
29 & VBG & Verb, gerund or present participle \\
30 & VBN & Verb, past participle \\
31 & VBP & Verb, non-3rd person singular present \\
32 & VBZ & Verb, 3rd person singular present \\
33 & WDT & Wh-determiner \\
34 & WP & Wh-pronoun \\
35 & WP\$ & Possessive wh-pronoun \\
36 & WRB & Wh-adverb \\
37 & \# & Pound sign \\
38 & \$ & Dollar sign \\
39 & . & Sentence-final punctuation \\
40 & , & Comma \\
41 & : & Colon, semi-colon \\
42 & ( & Left bracket character \\
43 & ) & Right bracket character \\
44 & " & Straight double quote \\
45 & ' & Left open single quote \\
46 & " & Left open double quote \\
47 & ' & Right close single quote \\
48 & " & Right close double quote \\ \bottomrule
\end{tabular}%
}
\caption{Penn Treebank POS tags.}
\label{tab:penn_treebank_pos_tags}
\end{table}
 
\begin{table*}[!tbh]
\centering
\scalebox{0.75}{
\setlength{\tabcolsep}{2.0pt}
\begin{tabular}{@{}llll@{}}
\toprule
\textbf{ANA   (anaphoric)} &  & \multicolumn{2}{l}{\textbf{MOD   (modality)}} \\ \midrule
PRO & anaphoric \& deictic pronouns: he, she, I, him & NOT & negation: not, no, neither, without \\
DEF & definite: the, loIT, derDE & NEC & necessity: must, should, have to \\
HAS & possessive pronoun: my, her & POS & possibility: might, could, perhaps, alleged, can \\
REF & reflexive \& reciprocal pron.: herself, each other & \multicolumn{2}{l}{\textbf{DSC (discourse)}} \\
EMP & emphasizing pronouns: himself & SUB & subordinate relations: that, while, because \\
\textbf{ACT (speech act)} &  & COO & coordinate relations: so, \{,\}, \{;\}, and \\
GRE & greeting \& parting: hi, bye & APP & appositional relations: \{,\}, which, \{(\}, — \\
ITJ & interjections, exclamations: alas, ah & BUT & contrast: but, yet \\
HES & hesitation: err & \multicolumn{2}{l}{\textbf{NAM (named entity)}} \\
QUE & interrogative: who, which, ? & PER & person: Axl Rose, Sherlock Holmes \\
\textbf{ATT (attribute)} &  & GPE & geo-political entity: Paris, Japan \\
QUC\* & concrete quantity: two, six million, twice & GPO\* & geo-political origin: Parisian, French \\
QUV\* & vague quantity: millions, many, enough & GEO & geographical location: Alps, Nile \\
COL\* & colour: red, crimson, light blue, chestnut brown & ORG & organization: IKEA, EU \\
IST & intersective: open, vegetarian, quickly & ART & artifact: iOS 7 \\
SST & subsective: skillful surgeon, tall kid & HAP & happening: Eurovision 2017 \\
PRI & privative: former, fake & UOM & unit of measurement: meter, \$, \%, degree Celsius \\
DEG\* & degree: 2 meters tall, 20 years old & CTC\* & contact information: 112, info@mail.com \\
INT & intensifier: very, much, too, rather & URL & URL: \url{http://pmb.let.rug.nl} \\
REL & relation: in, on, 's, of, after & LIT\* & literal use of names: his name is John \\
SCO & score: 3-0, grade A & NTH\* & other names: table 1a, equation (1) \\
\multicolumn{2}{l}{\textbf{COM   (comparative)}} & \multicolumn{2}{l}{\textbf{EVE (events)}} \\
EQU & equative: as tall as John, whales are mammals & EXS & untensed simple: to walk, is eaten, destruction \\
MOR & comparative positive: better, more & ENS & present simple: we walk, he walks \\
LES & comparative negative: less, worse & EPS & past simple: ate, went \\
TOP & superlative positive: most, mostly & EXG & untensed progressive: is running \\
BOT & superlative negative: worst, least & EXT & untensed perfect: has eaten \\
ORD & ordinal: 1st, 3rd, third & \multicolumn{2}{l}{\textbf{TNS (tense \& aspect)}} \\
\multicolumn{2}{l}{\textbf{UNE   (unnamed entity)}} & NOW & present tense: is skiing, do ski, has skied, now \\
CON & concept: dog, person & PST & past tense: was baked, had gone, did go \\
ROL & role: student, brother, prof., victim & FUT & future tense: will, shall \\
GRP\* & group: John \{,\} Mary and Sam gathered, a group of people & PRG\* & progressive: has been being treated, aan hetNL \\
\multicolumn{2}{l}{\textbf{DXS (deixis)}} & PFT\* & perfect: has been going/done \\
DXP\* & place deixis: here, this, above & \multicolumn{2}{l}{\textbf{TIM (temporal entity)}} \\
DXT\* & temporal deixis: just, later, tomorrow & DAT\* & full date: 27.04.2017, 27/04/17 \\
DXD\* & discourse deixis: latter, former, above & DOM & day of month: 27th December \\
\multicolumn{2}{l}{\textbf{LOG (logical)}} & YOC & year of century: 2017 \\
ALT & alternative \& repetitions: another, different, again & DOW & day of week: Thursday \\
XCL & exclusive: only, just & MOY & month of year: April \\
NIL & empty semantics: \{.\}, to, of & DEC & decade: 80s, 1990s \\
DIS & disjunction \& exist. quantif.: a, some, any, or & CLO & clocktime: 8:45 pm, 10 o'clock, noon \\
IMP & implication: if, when, unless &  &  \\
AND & \multicolumn{2}{l}{conjunction \& univ. quantif.:   every, and, who, any} &  \\ \bottomrule
\end{tabular}%
}
\caption{Semantic tags.}
\label{tab:sem-tags}
\end{table*}

\paragraph{Chunking tags:}
NP (Noun phrase), 
VP (Verb phrase), 
PP (Prepositional phrase), 
ADVP (Adverb phrase), 
SBAR (Subordinate phrase), 
ADJP (Adjective phrase), 
PRT (Particles), 
CONJP (Conjunction), 
INTJ (Interjection), 
LST (List marker), 
UCP (Unlike coordinate phrase). For the annotation, chunks are represented using IOB format, which results in 22 tags in the dataset as reported in Table \ref{tab:dataStats}.

\begin{table}[h]									
\centering					
\scalebox{1.0}{
\setlength{\tabcolsep}{2.5pt}
    \begin{tabular}{l|cccc|c}									
    \toprule									
Task    & Train & Dev & Test & Tags & F1\\		
\midrule
    POS & 36557 & 1802 & 1963 & 48 & 96.81\\
    SEM & 36928 & 5301 & 10600 & 73 & 96.32 \\
    Chunking &  8881 &  1843 &  2011 & 22 & 96.93 \\
    CCG &  39101 & 1908 & 2404 & 1272 & 95.24\\
    \bottomrule
    \end{tabular}
    }
    \caption{Data statistics (number of sentences) on training, development and test sets using in the experiments and the number of tags to be predicted}
\label{tab:dataStats}						
\end{table}

\subsection{BERT-based Sequence Tagger}
\label{sec:appendix:tagger}

We trained a BERT-based sequence tagger to auto-annotate our training data. We used standard splits for training, development and test data for the 4 linguistic tasks (POS, SEM, Chunking and CCG super tagging) that we used to carry out our analysis on. The splits to preprocess the data are available through git repository\footnote{\url{https://github.com/nelson-liu/contextual-repr-analysis}} released with \newcite{liu-etal-2019-linguistic}. See Table \ref{tab:dataStats} for statistics and classifier accuracy.

\section{Selection of the number of Clusters}
\label{sec:appendix:clustering}

We tried Elbow and Silhouette to get the optimum number of clusters, but did not observe any reliable patterns. In Elbow the distortion scores kept increasing, 
resulting in over-clustering (a large number of clusters consisted of less than 5 words). Over-clustering results in high but wrong alignment scores e.g. consider a two word cluster having words “bad” and “worse”. The cluster will have a successful match with “adjective” since more than 95\% of the words in the cluster are adjectives. In this way, a lot of small clusters will have a successful match with many human-defined concepts and the resulting alignment scores will be high. On the other hand, Silhouette resulted in under-clustering, giving the best score at number of clusters = 10.  We handled this empirically by trying several values for the number of clusters i.e., 200 to 1600 with step size 200. We selected 600 to find a good balance with over and under clustering. We understand that this may not be the best optimal point. We presented the results of 600 and 1000 clusters to show that our findings are not sensitive to the number of clusters parameter. Please See Figures \ref{fig:appendix:conceptPolarity-1000:sst} and \ref{fig:appendix:humanConcepts-pos} for comparison. 

\section{Analysis}

\subsection{Comparing Base and Fine-tuned Models}
\label{sec:appendix:base}

\begin{figure*}[t]
    \begin{subfigure}[b]{0.48\linewidth}
    \centering
    \includegraphics[width=\linewidth]{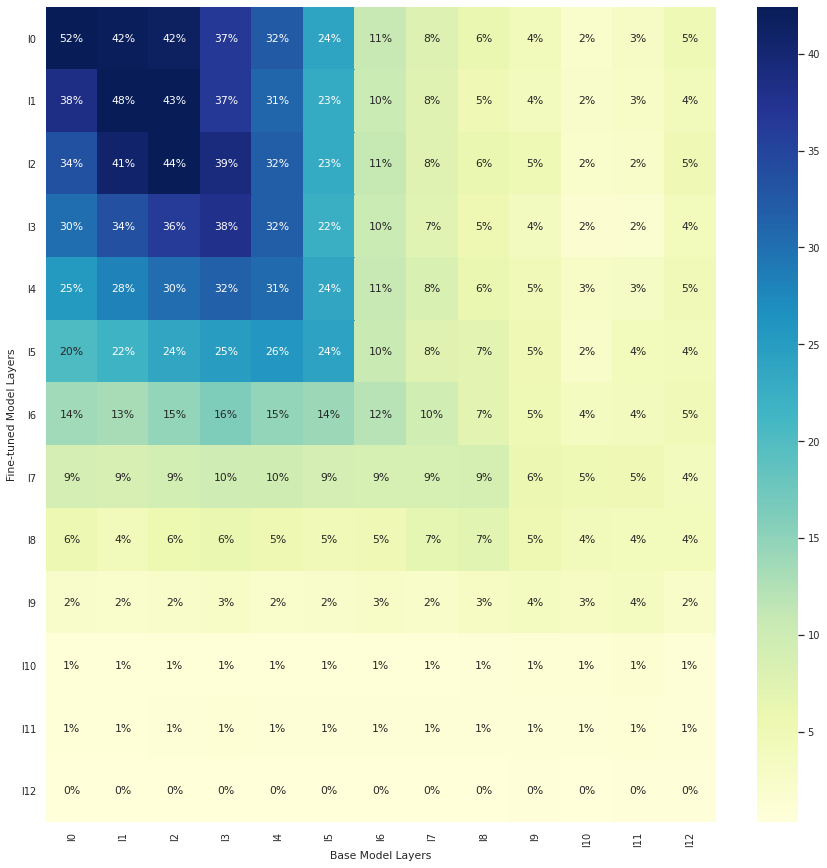}
    \caption{BERT (SST)}
    \label{fig:bert-sst-matrix-appendix}
    \end{subfigure}
    \begin{subfigure}[b]{0.48\linewidth}
    \centering
    \includegraphics[width=\linewidth]{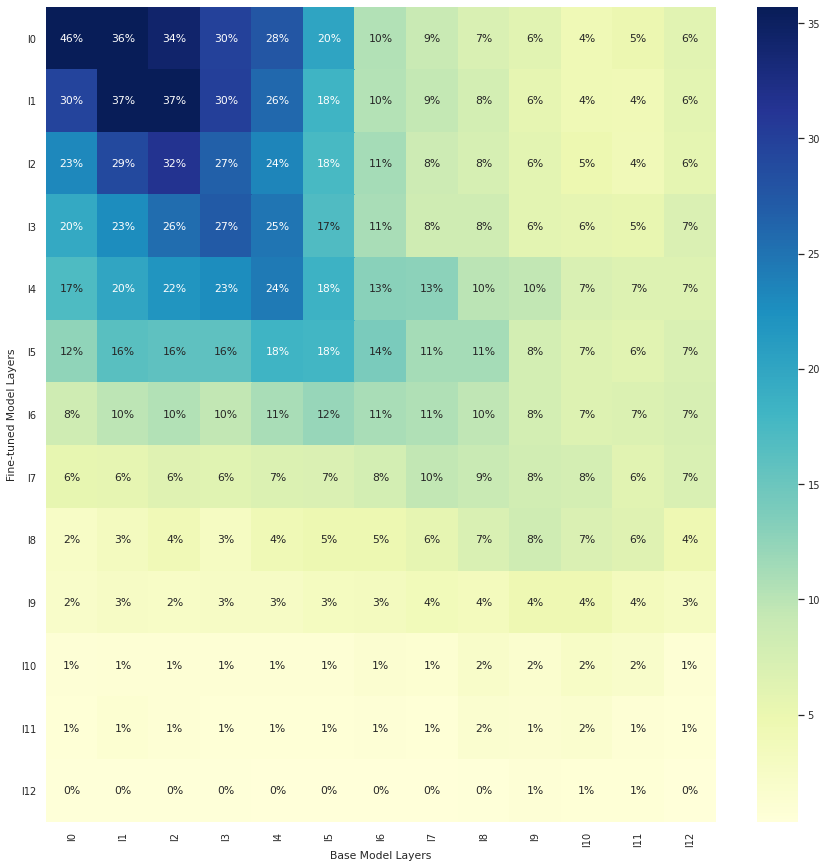}
    \caption{BERT (MNLI)}
    \label{fig:bert-mnli-matrix-appendix}
    \end{subfigure}
    \begin{subfigure}[b]{0.48\linewidth}
    \centering
    \includegraphics[width=\linewidth]{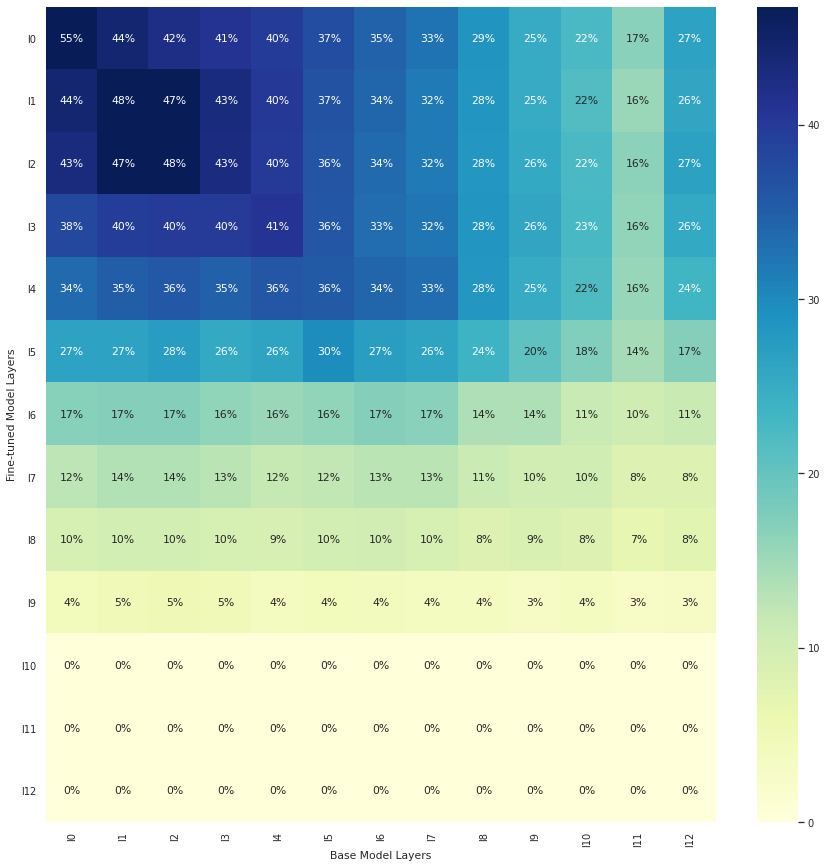}
    \caption{XLM-R (SST)}
    \label{fig:xlm-sst-matrix-appendix}
    \end{subfigure}
    \begin{subfigure}[b]{0.48\linewidth}
    \centering
    \includegraphics[width=\linewidth]{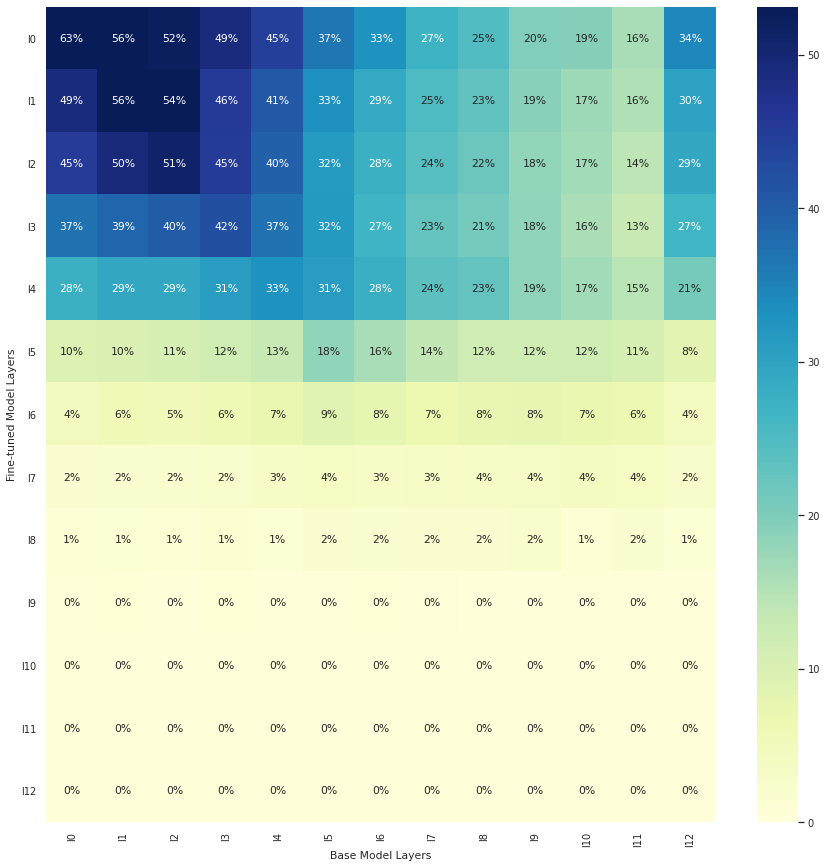}
    \caption{XLM-R (MNLI)}
    \label{fig:xlm-mnli-matrix-appendix}
    \end{subfigure}
    \begin{subfigure}[b]{0.48\linewidth}
    \centering
    \includegraphics[width=\linewidth]{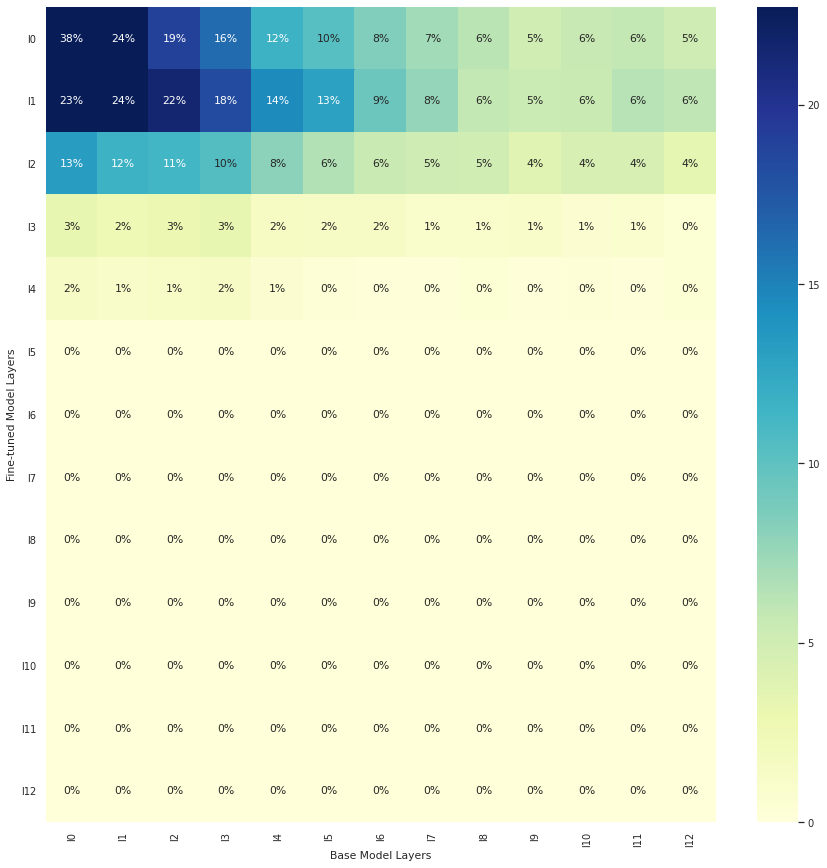}
    \caption{ALBERT (SST)}
    \label{fig:albert-sst-matrix-appendix}
    \end{subfigure}
    \begin{subfigure}[b]{0.48\linewidth}
    \centering
    \includegraphics[width=\linewidth]{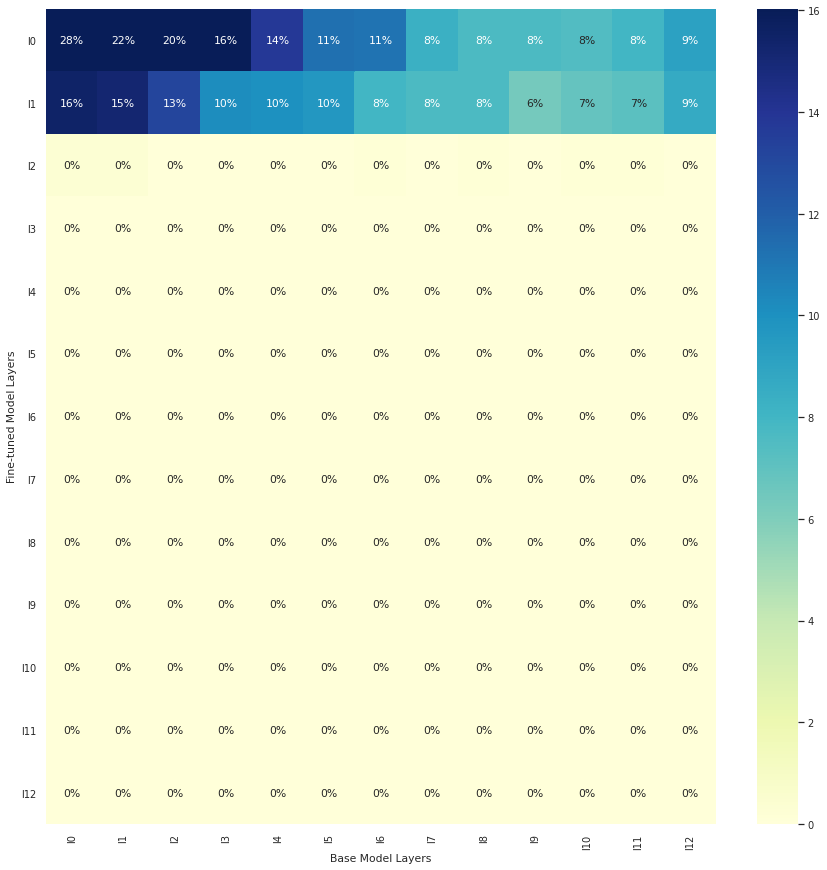}
    \caption{ALBERT (MNLI)}
    \label{fig:albert-mnli-matrix-appendix}
    \end{subfigure}
    
    \caption{Comparing Latent Concepts of Base models with their SST and MNLI fine-tuned versions. X-axis = base model, Y-axis = fine-tuned model}
    \label{fig:basevsft1-Appendix}
\end{figure*}

\begin{figure*}[t]
    \begin{subfigure}[b]{0.31\linewidth}
    \centering
    \includegraphics[width=\linewidth]{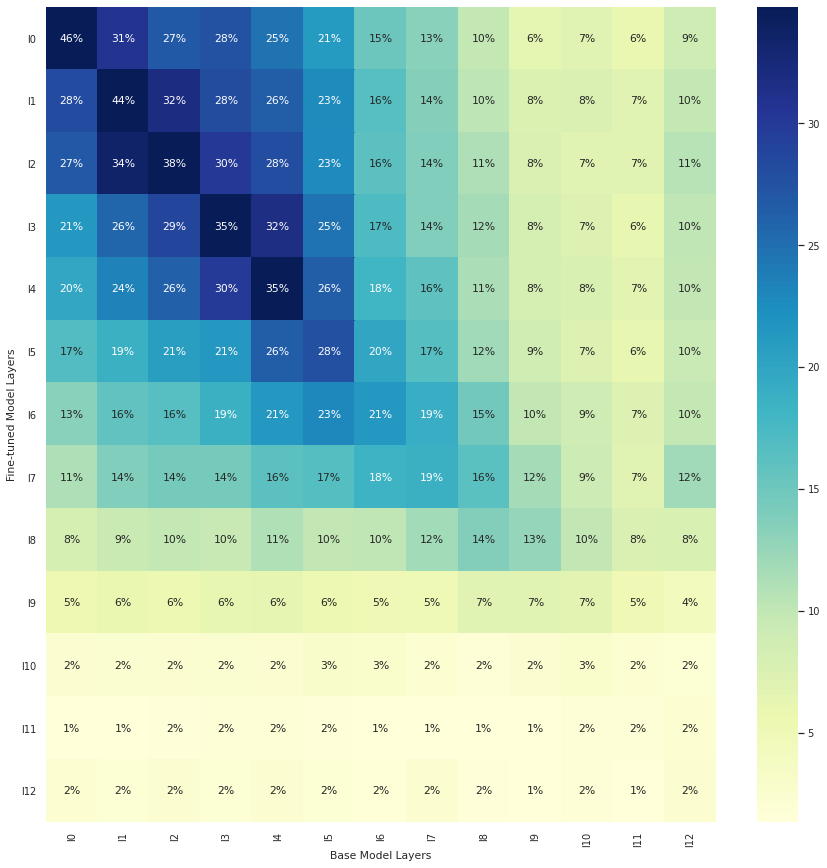}
     \caption{BERT (HS)}
    \label{fig:bert}
     \end{subfigure}
    \begin{subfigure}[b]{0.31\linewidth}
    \centering
    \includegraphics[width=\linewidth]{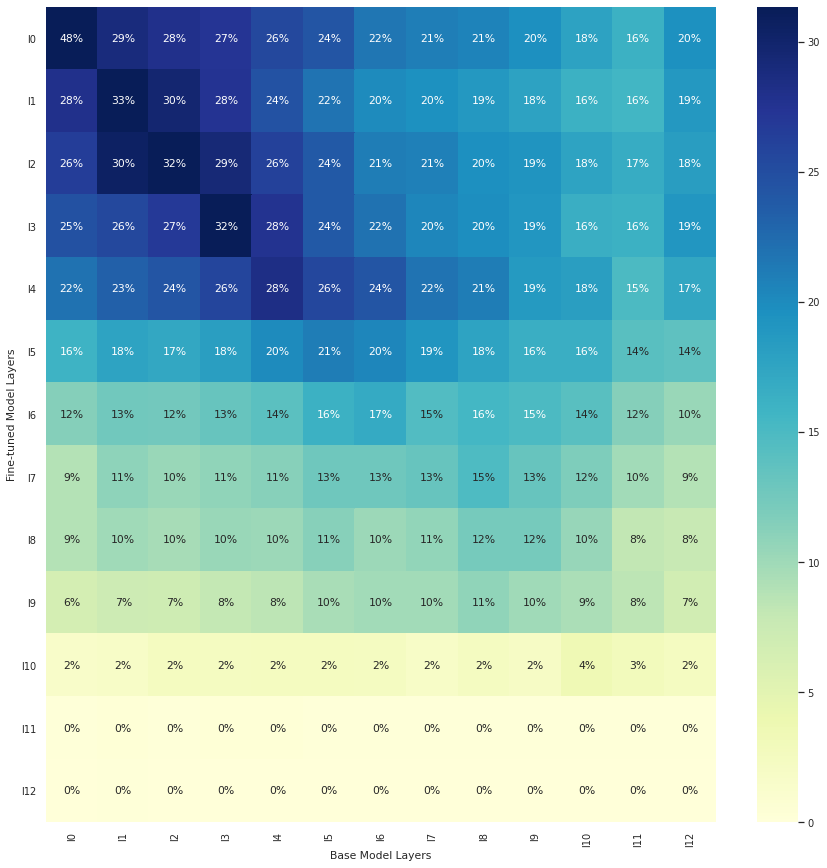}
    \caption{XLM-R (HS)}
    \label{fig:xlm-r}
    \end{subfigure}
    \begin{subfigure}[b]{0.31\linewidth}
    \centering
    \includegraphics[width=\linewidth]{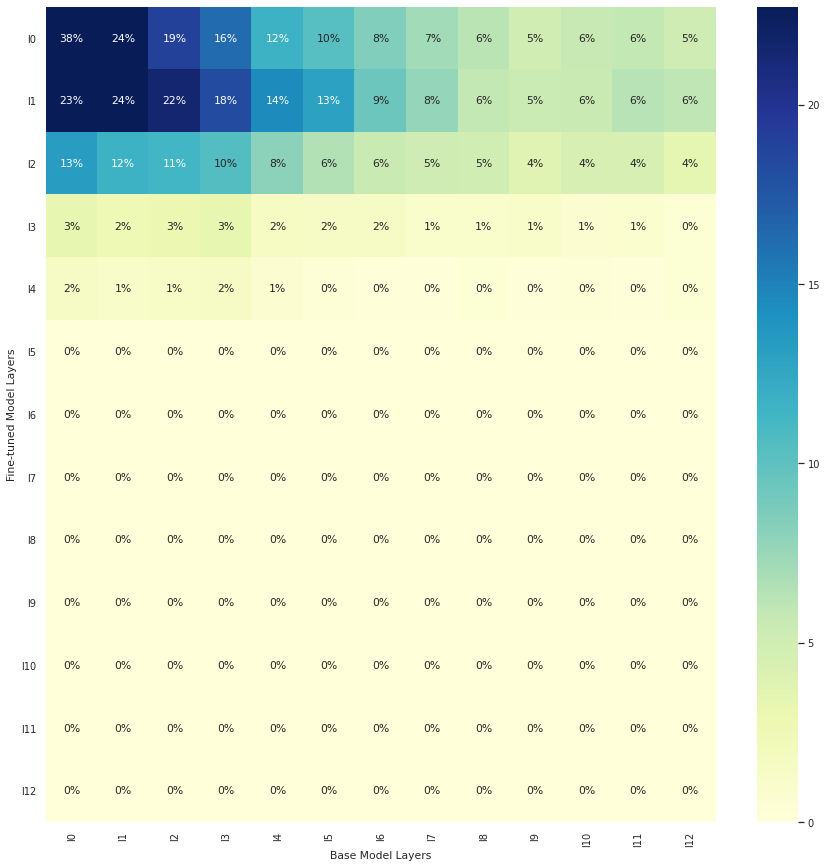}
    \caption{ALBERT (HS)}
    \label{fig:albert}
    \end{subfigure}
    \caption{Comparing Latent Concepts of Base models with their Hate Speech fine-tuned versions. X-axis = base model, Y-axis = fine-tuned model}
    \label{fig:basevsft2-Appendix}
\end{figure*}

\begin{figure*}[t]
    \begin{subfigure}[b]{0.33\linewidth}
    \centering
    \includegraphics[width=\linewidth]{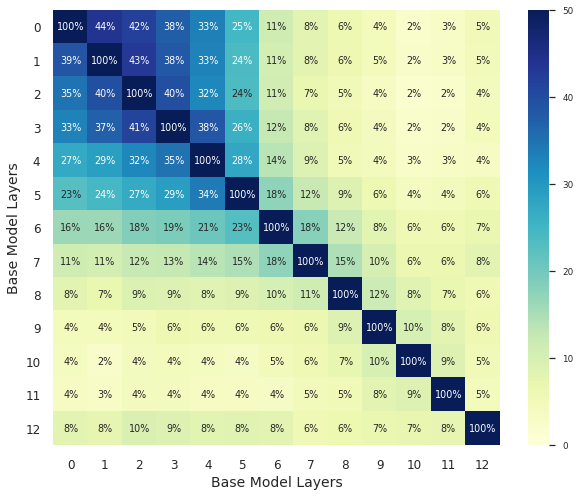}
     \caption{BERT (SST)}
    \label{fig:bert-base-base-sst}
     \end{subfigure}
    \begin{subfigure}[b]{0.33\linewidth}
    \centering
    \includegraphics[width=\linewidth]{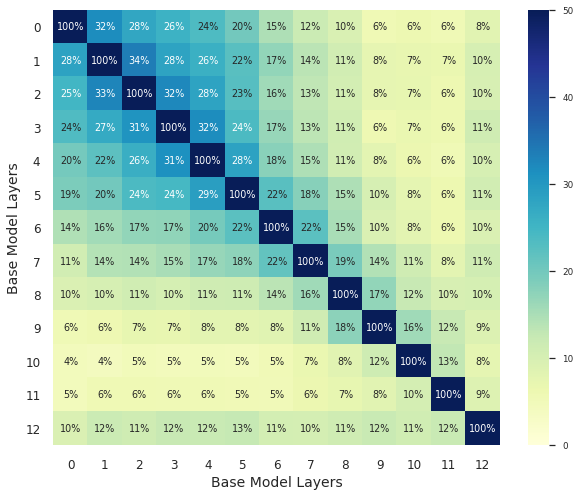}
    \caption{XLM-R(SST)}
    \label{fig:bert-base-base-hsd}
    \end{subfigure}
    \begin{subfigure}[b]{0.33\linewidth}
    \centering
    \includegraphics[width=\linewidth]{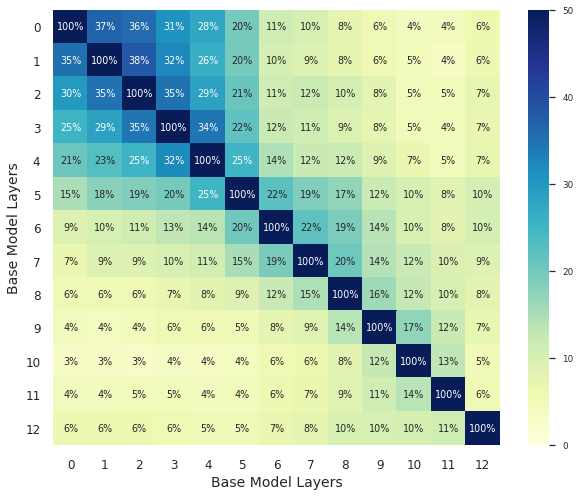}
    \caption{ALBERT (SST)}
    \label{fig:bert-base-base-mnli}
    \end{subfigure}
    \caption{Comparing Latent Concepts of Base models with themselves. X-axis = base model, Y-axis = fine-tuned model}
    \label{fig:basevsbase-Appendix}
\end{figure*}

In Section \ref{subsec:base} we showed the overlap between the encoded concepts of base and fine-tuned SST models. In Figures \ref{fig:basevsft1-Appendix} and \ref{fig:basevsft2-Appendix} we show the same for MNLI and Hate Speech models. We also report the results on how the concepts evolve across the layers. We found that lower layers of the model (until layer 5) show a substantial overlap (up to 40\% overlapping clusters). Higher layers show less than 10\% overlap. Please See Figure \ref{fig:basevsbase-Appendix} for this result.

\subsection{Presence of Linguistic Concepts in the Latent Space}
\label{subsec:appendix:linguistic}

\begin{figure*}[t]
    \begin{subfigure}[b]{0.31\linewidth}
    \centering
    \includegraphics[width=\linewidth]{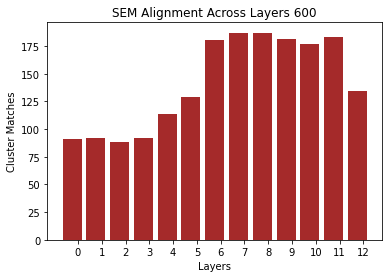}
    \caption{BERT -- Base}
    \label{fig:bert}
    \end{subfigure}
    \begin{subfigure}[b]{0.31\linewidth}
    \centering
    \includegraphics[width=\linewidth]{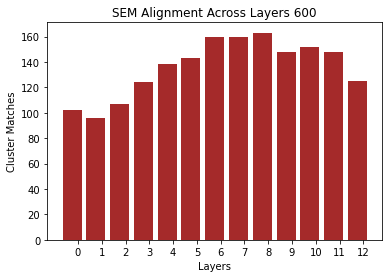}
    \caption{XLM-R -- Base}
    \label{fig:xlm-r}
    \end{subfigure}
    \begin{subfigure}[b]{0.31\linewidth}
    \centering
    \includegraphics[width=\linewidth]{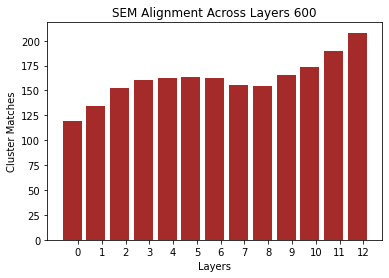}
    \caption{ALBERT -- Base}
    \label{fig:albert}
    \end{subfigure}
    \begin{subfigure}[b]{0.31\linewidth}
    \centering
    \includegraphics[width=\linewidth]{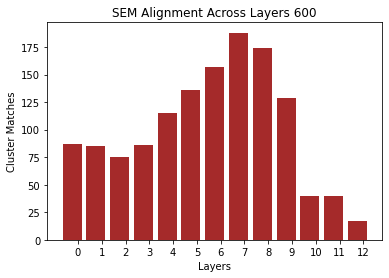}
    \caption{BERT -- SST}
    \label{fig:bert}
    \end{subfigure}
    \begin{subfigure}[b]{0.31\linewidth}
    \centering
    \includegraphics[width=\linewidth]{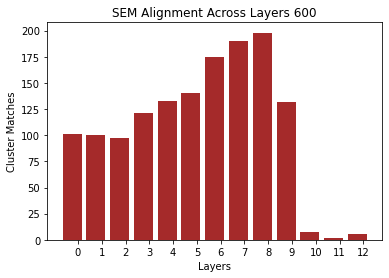}
    \caption{XLM-R -- SST}
    \label{fig:xlm-r}
    \end{subfigure}
    \begin{subfigure}[b]{0.31\linewidth}
    \centering
    \includegraphics[width=\linewidth]{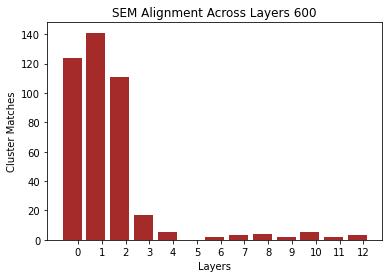}
    \caption{ALBERT -- SST}
    \label{fig:albert}
    \end{subfigure}
    \caption{Aligning encoded concepts with human-defined concept (SEM) in base and pre-trained models}
    \label{fig:humanConcepts-sem}
\end{figure*}

\begin{figure*}[t]
    \begin{subfigure}[b]{0.31\linewidth}
    \centering
    \includegraphics[width=\linewidth]{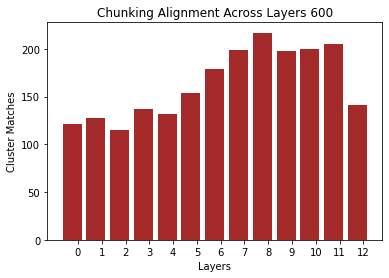}
    \caption{BERT -- Base}
    \label{fig:bert}
    \end{subfigure}
    \begin{subfigure}[b]{0.31\linewidth}
    \centering
    \includegraphics[width=\linewidth]{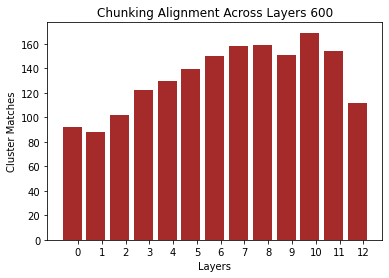}
    \caption{XLM-R -- Base}
    \label{fig:xlm-r}
    \end{subfigure}
    \begin{subfigure}[b]{0.31\linewidth}
    \centering
    \includegraphics[width=\linewidth]{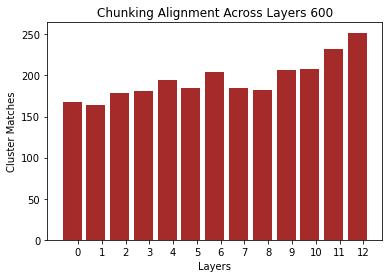}
    \caption{ALBERT -- Base}
    \label{fig:albert}
    \end{subfigure}
    \begin{subfigure}[b]{0.31\linewidth}
    \centering
    \includegraphics[width=\linewidth]{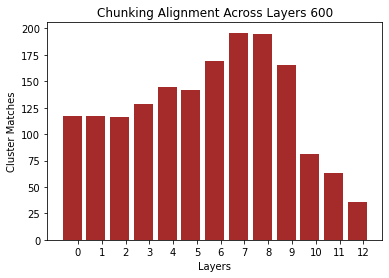}
    \caption{BERT -- SST}
    \label{fig:bert}
    \end{subfigure}
    \begin{subfigure}[b]{0.31\linewidth}
    \centering
    \includegraphics[width=\linewidth]{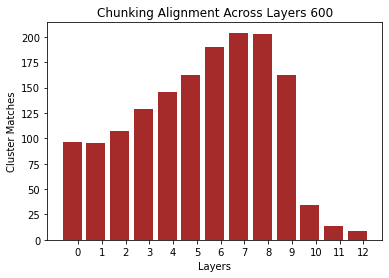}
    \caption{XLM-R -- SST}
    \label{fig:xlm-r}
    \end{subfigure}
    \begin{subfigure}[b]{0.31\linewidth}
    \centering
    \includegraphics[width=\linewidth]{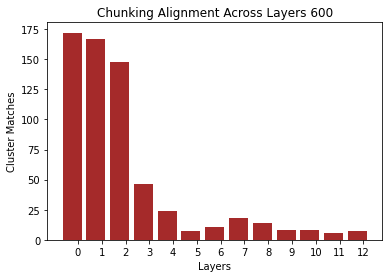}
    \caption{ALBERT -- SST}
    \label{fig:albert}
    \end{subfigure}
    \caption{Aligning encoded concepts with human-defined concept (Chunking) in base and pre-trained models}
    \label{fig:humanConcepts-chunking}
\end{figure*}

\begin{figure*}[t]
    \begin{subfigure}[b]{0.31\linewidth}
    \centering
    \includegraphics[width=\linewidth]{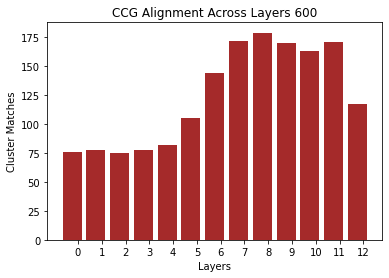}
    \caption{BERT -- Base}
    \label{fig:bert}
    \end{subfigure}
    \begin{subfigure}[b]{0.31\linewidth}
    \centering
    \includegraphics[width=\linewidth]{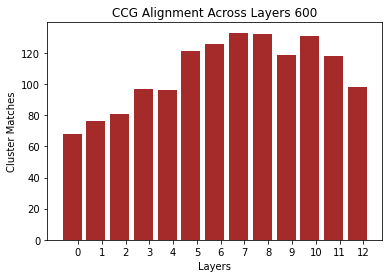}
    \caption{XLM-R -- Base}
    \label{fig:xlm-r}
    \end{subfigure}
    \begin{subfigure}[b]{0.31\linewidth}
    \centering
    \includegraphics[width=\linewidth]{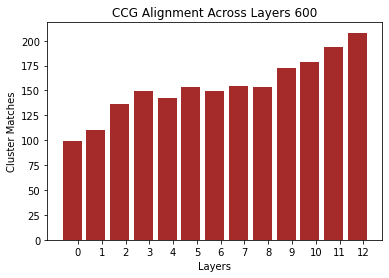}
    \caption{ALBERT -- Base}
    \label{fig:albert}
    \end{subfigure}
    \begin{subfigure}[b]{0.31\linewidth}
    \centering
    \includegraphics[width=\linewidth]{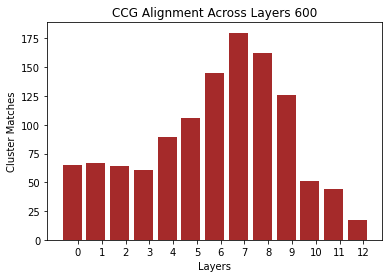}
    \caption{BERT -- SST}
    \label{fig:bert}
    \end{subfigure}
    \begin{subfigure}[b]{0.31\linewidth}
    \centering
    \includegraphics[width=\linewidth]{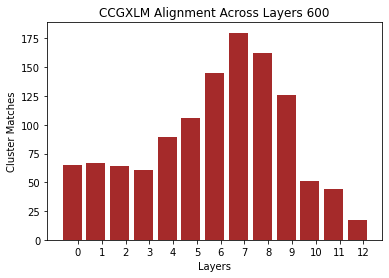}
    \caption{XLM-R -- SST}
    \label{fig:xlm-r}
    \end{subfigure}
    \begin{subfigure}[b]{0.31\linewidth}
    \centering
    \includegraphics[width=\linewidth]{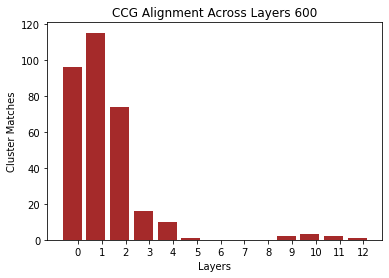}
    \caption{ALBERT -- SST}
    \label{fig:albert}
    \end{subfigure}
    \caption{Aligning encoded concepts with human-defined concept (CCG) in base and pre-trained models}
    \label{fig:humanConcepts-ccg}
\end{figure*}

\begin{figure*}[t]
    \begin{subfigure}[b]{0.31\linewidth}
    \centering
    \includegraphics[width=\linewidth]{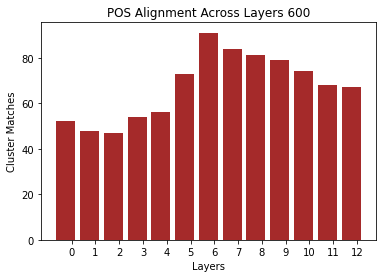}
    \caption{BERT -- Base}
    \label{fig:bert}
    \end{subfigure}
    \begin{subfigure}[b]{0.31\linewidth}
    \centering
    \includegraphics[width=\linewidth]{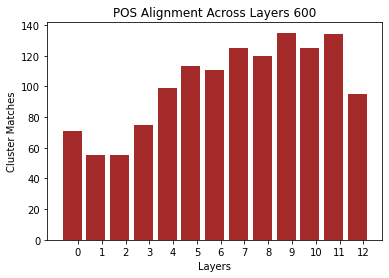}
    \caption{XLM-R -- Base}
    \label{fig:xlm-r}
    \end{subfigure}
    \begin{subfigure}[b]{0.31\linewidth}
    \centering
    \includegraphics[width=\linewidth]{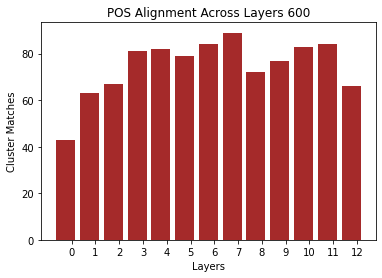}
    \caption{ALBERT -- Base}
    \label{fig:albert}
    \end{subfigure}
    \begin{subfigure}[b]{0.31\linewidth}
    \centering
    \includegraphics[width=\linewidth]{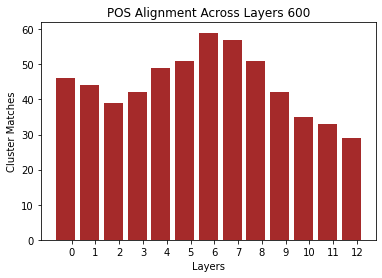}
    \caption{BERT -- MNLI}
    \label{fig:bert}
    \end{subfigure}
    \begin{subfigure}[b]{0.31\linewidth}
    \centering
    \includegraphics[width=\linewidth]{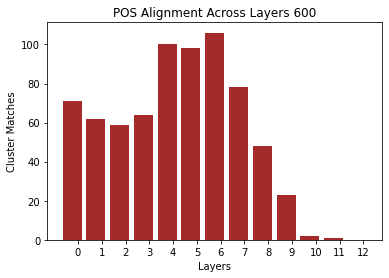}
    \caption{XLM-R -- MNLI}
    \label{fig:xlm-r}
    \end{subfigure}
    \begin{subfigure}[b]{0.31\linewidth}
    \centering
    \includegraphics[width=\linewidth]{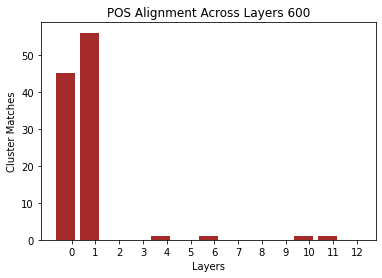}
    \caption{ALBERT -- MNLI}
    \label{fig:albert}
    \end{subfigure}
    \caption{Aligning encoded concepts with human-defined concept (POS
    ) in base and pre-trained models}
    \label{fig:humanConcepts-mnli-pos}
\end{figure*}

\begin{figure*}[t]
    \begin{subfigure}[b]{0.31\linewidth}
    \centering
    \includegraphics[width=\linewidth]{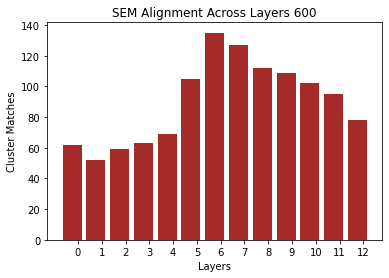}
    \caption{BERT -- Base}
    \label{fig:bert}
    \end{subfigure}
    \begin{subfigure}[b]{0.31\linewidth}
    \centering
    \includegraphics[width=\linewidth]{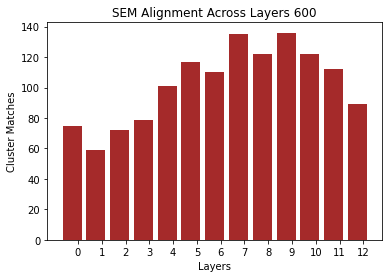}
    \caption{XLM-R -- Base}
    \label{fig:xlm-r}
    \end{subfigure}
    \begin{subfigure}[b]{0.31\linewidth}
    \centering
    \includegraphics[width=\linewidth]{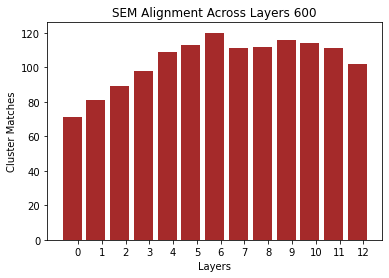}
    \caption{ALBERT -- Base}
    \label{fig:albert}
    \end{subfigure}
    \begin{subfigure}[b]{0.31\linewidth}
    \centering
    \includegraphics[width=\linewidth]{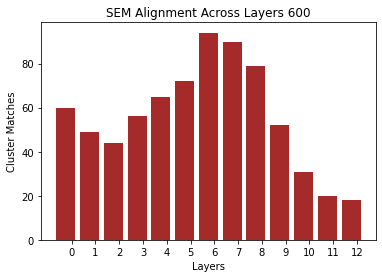}
    \caption{BERT -- MNLI}
    \label{fig:bert}
    \end{subfigure}
    \begin{subfigure}[b]{0.31\linewidth}
    \centering
    \includegraphics[width=\linewidth]{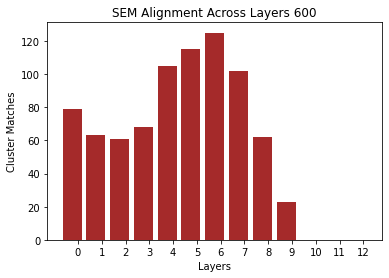}
    \caption{XLM-R -- MNLI}
    \label{fig:xlm-r}
    \end{subfigure}
    \begin{subfigure}[b]{0.31\linewidth}
    \centering
    \includegraphics[width=\linewidth]{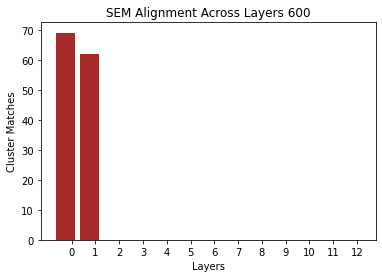}
    \caption{ALBERT -- MNLI}
    \label{fig:albert}
    \end{subfigure}
    \caption{Aligning encoded concepts with human-defined concept (SEM
    ) in base and pre-trained models}
    \label{fig:humanConcepts-mnli-SEM}
\end{figure*}

\begin{figure*}[t]
    \begin{subfigure}[b]{0.31\linewidth}
    \centering
    \includegraphics[width=\linewidth]{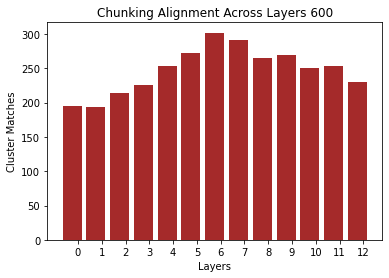}
    \caption{BERT -- Base}
    \label{fig:bert}
    \end{subfigure}
    \begin{subfigure}[b]{0.31\linewidth}
    \centering
    \includegraphics[width=\linewidth]{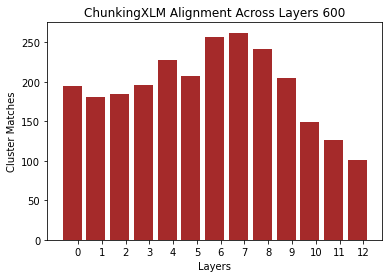}
    \caption{XLM-R -- Base}
    \label{fig:xlm-r}
    \end{subfigure}
    \begin{subfigure}[b]{0.31\linewidth}
    \centering
    \includegraphics[width=\linewidth]{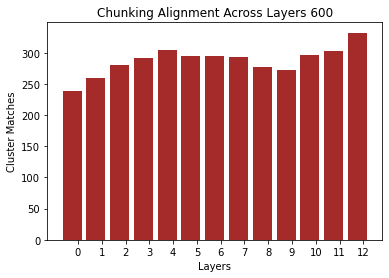}
    \caption{ALBERT -- Base}
    \label{fig:albert}
    \end{subfigure}
    \begin{subfigure}[b]{0.31\linewidth}
    \centering
    \includegraphics[width=\linewidth]{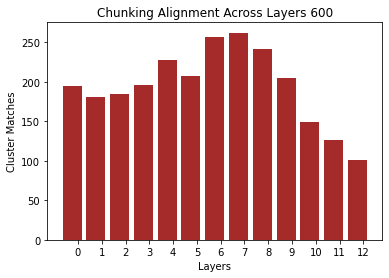}
    \caption{BERT -- MNLI}
    \label{fig:bert}
    \end{subfigure}
    \begin{subfigure}[b]{0.31\linewidth}
    \centering
    \includegraphics[width=\linewidth]{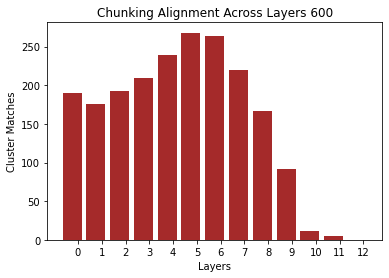}
    \caption{XLM-R -- MNLI}
    \label{fig:xlm-r}
    \end{subfigure}
    \begin{subfigure}[b]{0.31\linewidth}
    \centering
    \includegraphics[width=\linewidth]{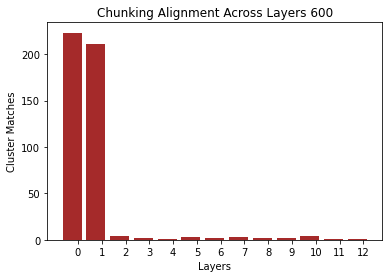}
    \caption{ALBERT -- MNLI}
    \label{fig:albert}
    \end{subfigure}
    \caption{Aligning encoded concepts with human-defined concept (CHUNKING
    ) in base and pre-trained models}
    \label{fig:humanConcepts-mnli-CHUNKING}
\end{figure*}

\begin{figure*}[t]
    \begin{subfigure}[b]{0.31\linewidth}
    \centering
    \includegraphics[width=\linewidth]{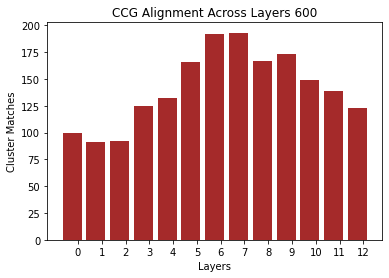}
    \caption{BERT -- Base}
    \label{fig:bert}
    \end{subfigure}
    \begin{subfigure}[b]{0.31\linewidth}
    \centering
    \includegraphics[width=\linewidth]{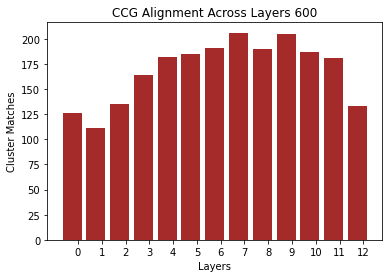}
    \caption{XLM-R -- Base}
    \label{fig:xlm-r}
    \end{subfigure}
    \begin{subfigure}[b]{0.31\linewidth}
    \centering
    \includegraphics[width=\linewidth]{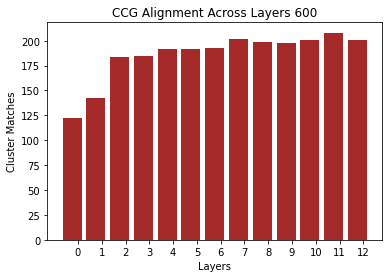}
    \caption{ALBERT -- Base}
    \label{fig:albert}
    \end{subfigure}
    \begin{subfigure}[b]{0.31\linewidth}
    \centering
    \includegraphics[width=\linewidth]{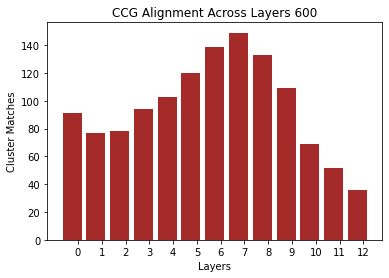}
    \caption{BERT -- MNLI}
    \label{fig:bert}
    \end{subfigure}
    \begin{subfigure}[b]{0.31\linewidth}
    \centering
    \includegraphics[width=\linewidth]{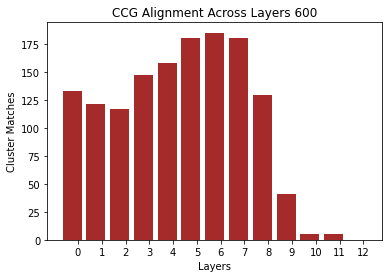}
    \caption{XLM-R -- MNLI}
    \label{fig:xlm-r}
    \end{subfigure}
    \begin{subfigure}[b]{0.31\linewidth}
    \centering
    \includegraphics[width=\linewidth]{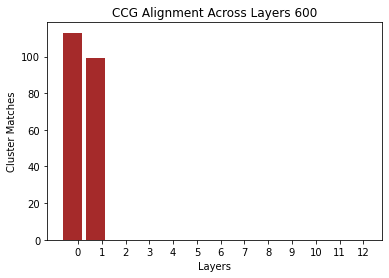}
    \caption{ALBERT -- MNLI}
    \label{fig:albert}
    \end{subfigure}
    \caption{Aligning encoded concepts with human-defined concept (CCG
    ) in base and pre-trained models}
    \label{fig:humanConcepts-mnli-CCG}
\end{figure*}

In Section \ref{subsec:linguistic} we showed the overlap of the encoded concepts in the base and fine-tuned SST models with human-defined POS concepts. In Figures \ref{fig:humanConcepts-sem}-\ref{fig:humanConcepts-mnli-CCG}, we provide alignment results for SEM, CCG and Chunking concepts with SST and also MNLI tasks.

\subsection{Task-specific Latent Spaces}
\label{sec:appendix:taskLabels}

\begin{figure*}[t]
    \begin{subfigure}[b]{0.31\linewidth}
    \centering
    \includegraphics[width=\linewidth]{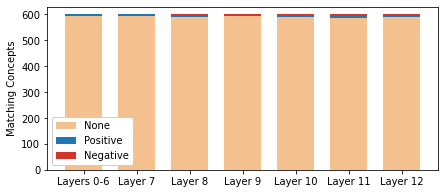}
    \caption{BERT -- Base}
    \label{fig:bert}
    \end{subfigure}
    \begin{subfigure}[b]{0.31\linewidth}
    \centering
    \includegraphics[width=\linewidth]{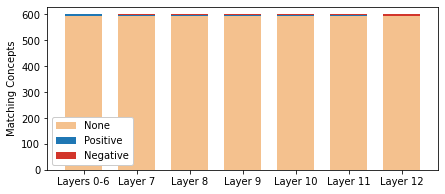}
    \caption{XLM-R -- Base}
    \label{fig:xlm-r}
    \end{subfigure}
    \begin{subfigure}[b]{0.31\linewidth}
    \centering
    \includegraphics[width=\linewidth]{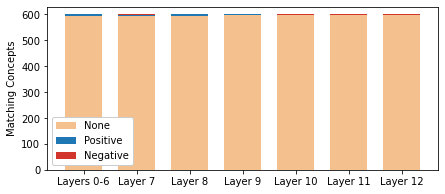}
    \caption{ALBERT -- Base}
    \label{fig:albert}
    \end{subfigure}
    \begin{subfigure}[b]{0.33\linewidth}
    \centering
    \includegraphics[width=\linewidth]{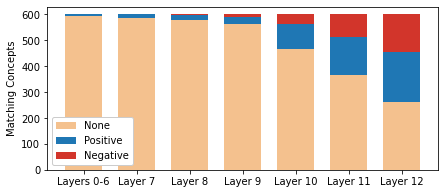}
    \caption{BERT -- HS}
    \label{fig:bert}
    \end{subfigure}
    \begin{subfigure}[b]{0.33\linewidth}
    \centering
    \includegraphics[width=\linewidth]{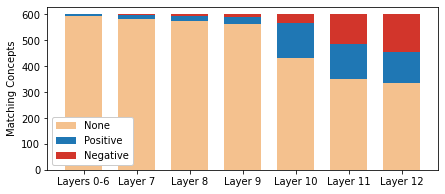}
    \caption{XLM-R -- HS}
    \label{fig:xlm-r}
    \end{subfigure}
    \begin{subfigure}[b]{0.33\linewidth}
    \centering
    \includegraphics[width=\linewidth]{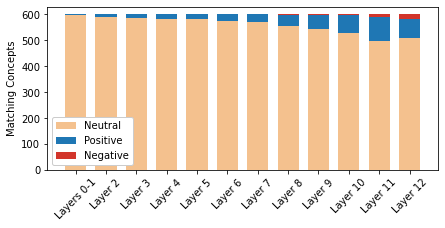}
    \caption{ALBERT -- HS}
    \label{fig:albert}
    \end{subfigure}
    \caption{Aligning encoded concepts with the task specific (Hate Speech: toxic vs. non-toxic) concepts. Positive = Toxic, Negative = Non-Toxic}
    \label{fig:appendix:conceptPolarity:hs}
\end{figure*}

In Section \ref{subsec:taskspecific} we studied how the concepts in SST models acquire polarity towards the task. We did not show the base models due to space limitations. Here we show the base models as well to demonstrate that all concepts had no polarity in the base models. In Figure \ref{fig:appendix:conceptPolarity:hs}, we show the same for the Hate-Speech task. We do not show the MNLI task, because we could not find polarity concepts in that task.

\begin{figure*}[t]
    \begin{subfigure}[b]{0.31\linewidth}
    \centering
    \includegraphics[width=\linewidth]{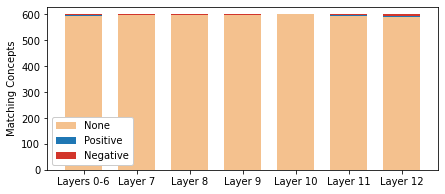}
    \caption{BERT -- Base}
    \label{fig:bert}
    \end{subfigure}
    \begin{subfigure}[b]{0.31\linewidth}
    \centering
    \includegraphics[width=\linewidth]{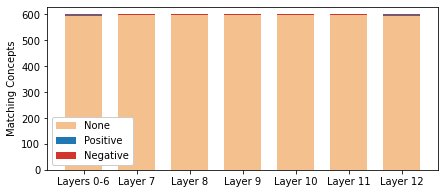}
    \caption{XLM-R -- Base}
    \label{fig:xlm-r}
    \end{subfigure}
    \begin{subfigure}[b]{0.31\linewidth}
    \centering
    \includegraphics[width=\linewidth]{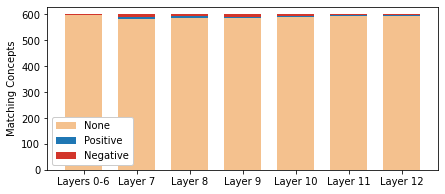}
    \caption{ALBERT -- Base}
    \label{fig:albert}
    \end{subfigure}
    \begin{subfigure}[b]{0.33\linewidth}
    \centering
    \includegraphics[width=\linewidth]{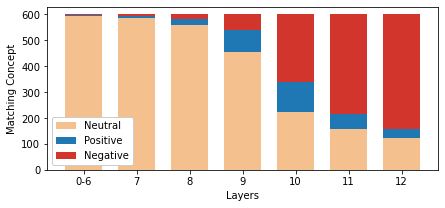}
    \caption{BERT -- SST}
    \label{fig:bert}
    \end{subfigure}
    \begin{subfigure}[b]{0.33\linewidth}
    \centering
    \includegraphics[width=\linewidth]{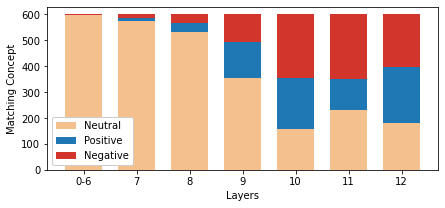}
    \caption{XLM-R -- SST}
    \label{fig:xlm-r}
    \end{subfigure}
    \begin{subfigure}[b]{0.33\linewidth}
    \centering
    \includegraphics[width=\linewidth]{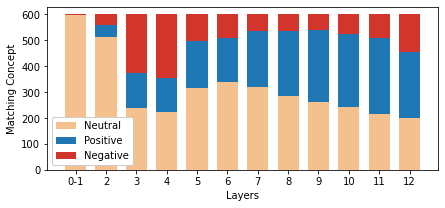}
    \caption{ALBERT -- SST}
    \label{fig:albert}
    \end{subfigure}
    \caption{Aligning encoded concepts with the task specific concepts}
    \label{fig:appendix:conceptPolarity:sst}
\end{figure*}

\section{Selection of task-specific Latent clusters}
\label{sec:appendix:sample_clusters}

\begin{figure*}[t]
    \begin{subfigure}[b]{0.30\linewidth}
\centering
\includegraphics[width=\linewidth]{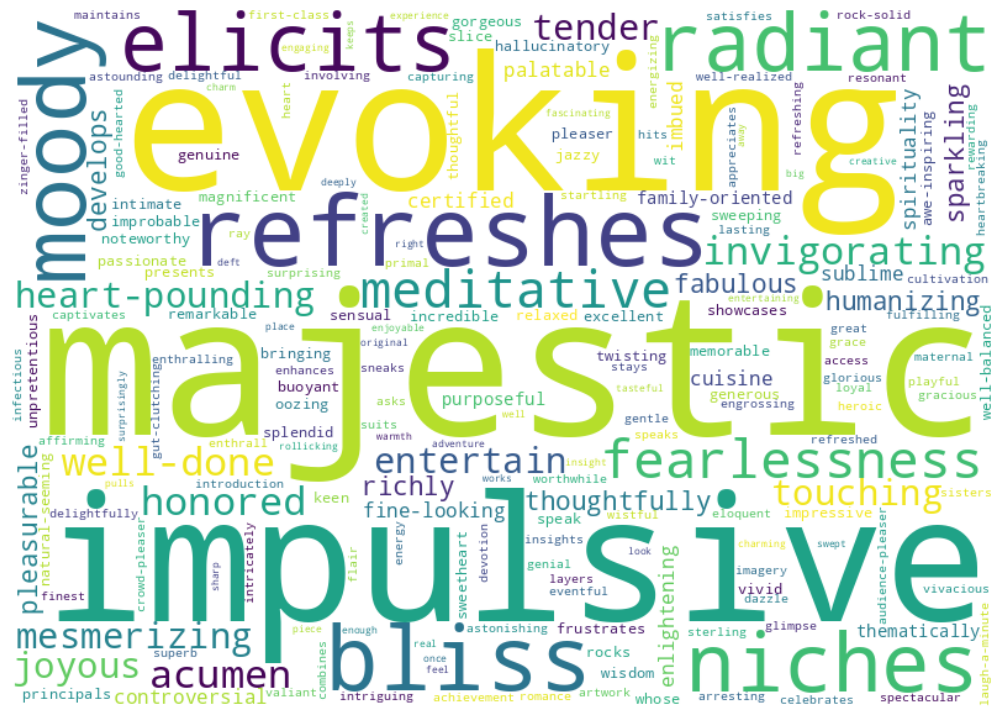}
\caption{XLM-R Layer 12 Cluster 470 (Positive Sentiment)}
\end{subfigure}
\begin{subfigure}[b]{0.30\linewidth}
\centering
\includegraphics[width=\linewidth]{Figures/selected-word-clouds/xlm-sst-layer12-c15-0.319280205655527.png}
\caption{XLM-R Layer 12 Cluster 15 (Positive Sentiment)}
\end{subfigure}
\begin{subfigure}[b]{0.30\linewidth}
\centering
\includegraphics[width=\linewidth]{Figures/selected-word-clouds/xlm-sst-layer10-c16-0.4082725060827251.png}
\caption{XLM-R Layer 10 Cluster 16 (Negative Sentiment)}
\end{subfigure}
\begin{subfigure}[b]{0.30\linewidth}
\centering
\includegraphics[width=\linewidth]{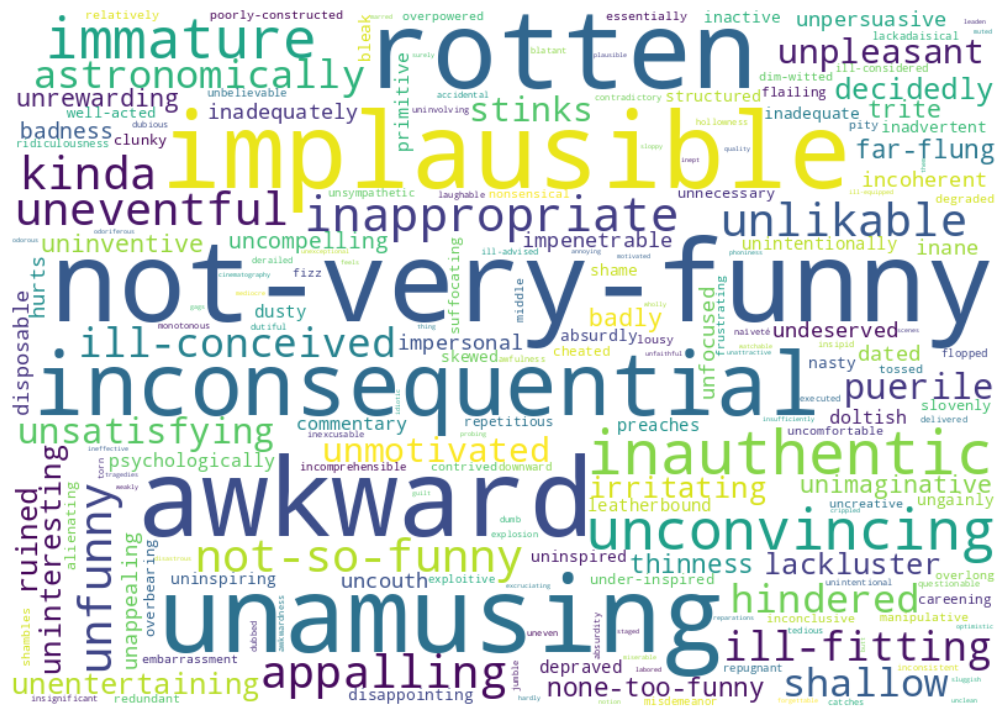}
\caption{XLM-R Layer 10 Cluster 121 (Negative Sentiment)}
\end{subfigure}
\begin{subfigure}[b]{0.30\linewidth}
\centering
\includegraphics[width=\linewidth]{Figures/selected-word-clouds/xlm-hs-layer10-c567-0.9510280373831775.png}
\caption{XLM-R Layer 10 Cluster 576 (Toxic Hatespeech)}
\end{subfigure}
\begin{subfigure}[b]{0.30\linewidth}
\centering
\includegraphics[width=\linewidth]{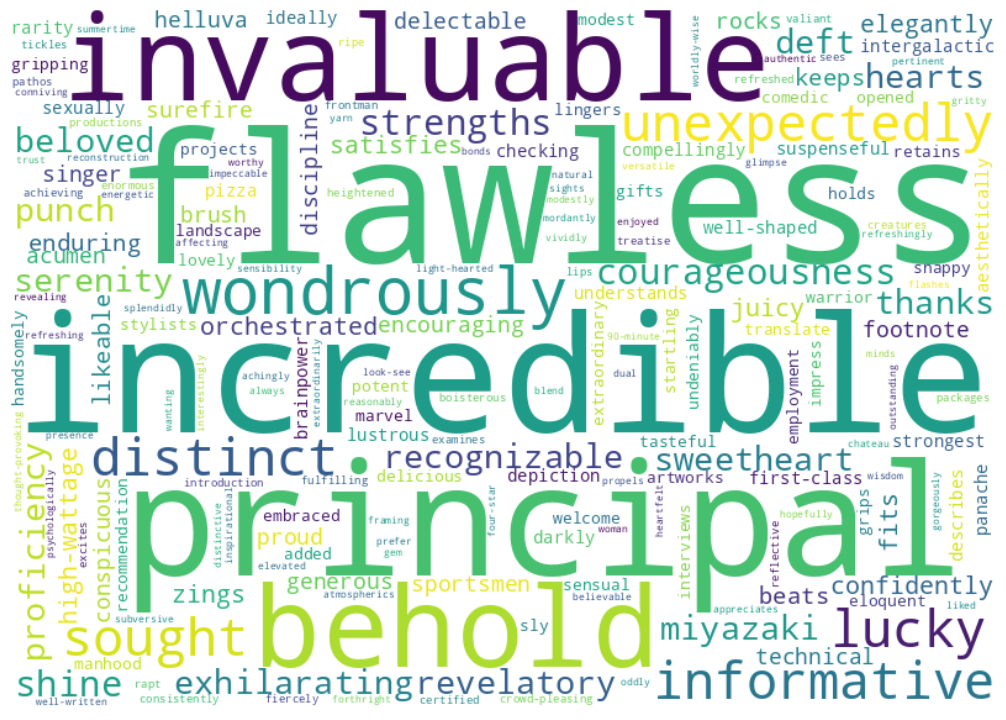}
\caption{BERT Layer 12 Cluster 432 (Positive Sentiment)}
\end{subfigure}
\begin{subfigure}[b]{0.30\linewidth}
\centering
\includegraphics[width=\linewidth]{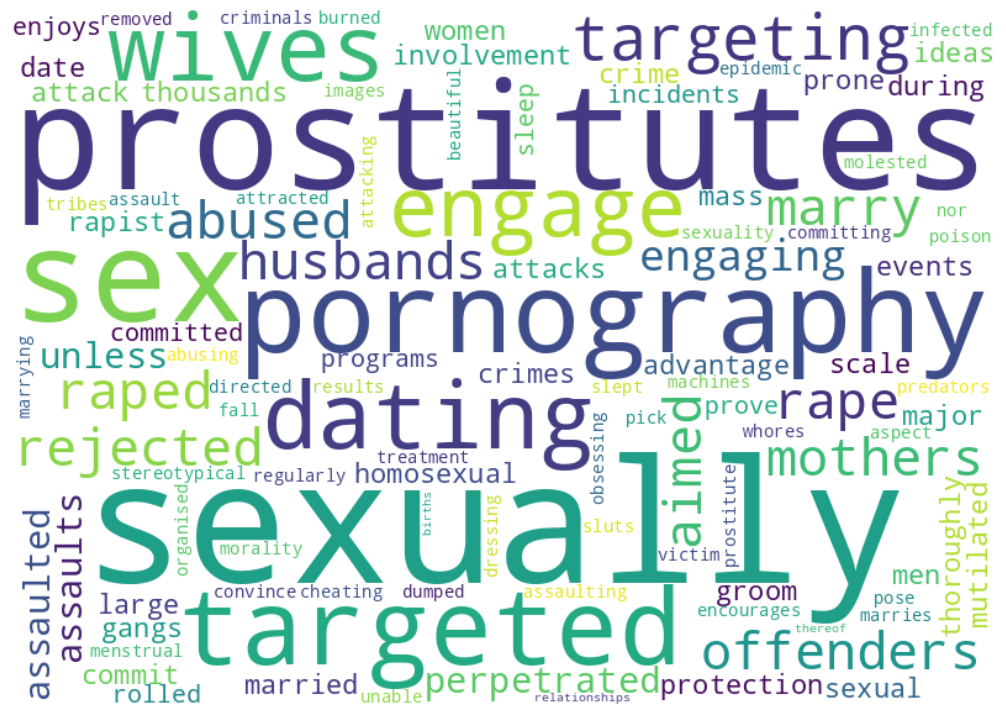}
\caption{BERT Layer 12 Cluster 272 (Toxic Hatespeech)}
\end{subfigure}
\begin{subfigure}[b]{0.30\linewidth}
\centering
\includegraphics[width=\linewidth]{Figures/selected-word-clouds/bert-base-cased-hs-layer10-c227-0.8927175843694494.png}
\caption{BERT Layer 10 Cluster 227 (Toxic Hatespeech)}
\end{subfigure}
\begin{subfigure}[b]{0.30\linewidth}
\centering
\includegraphics[width=\linewidth]{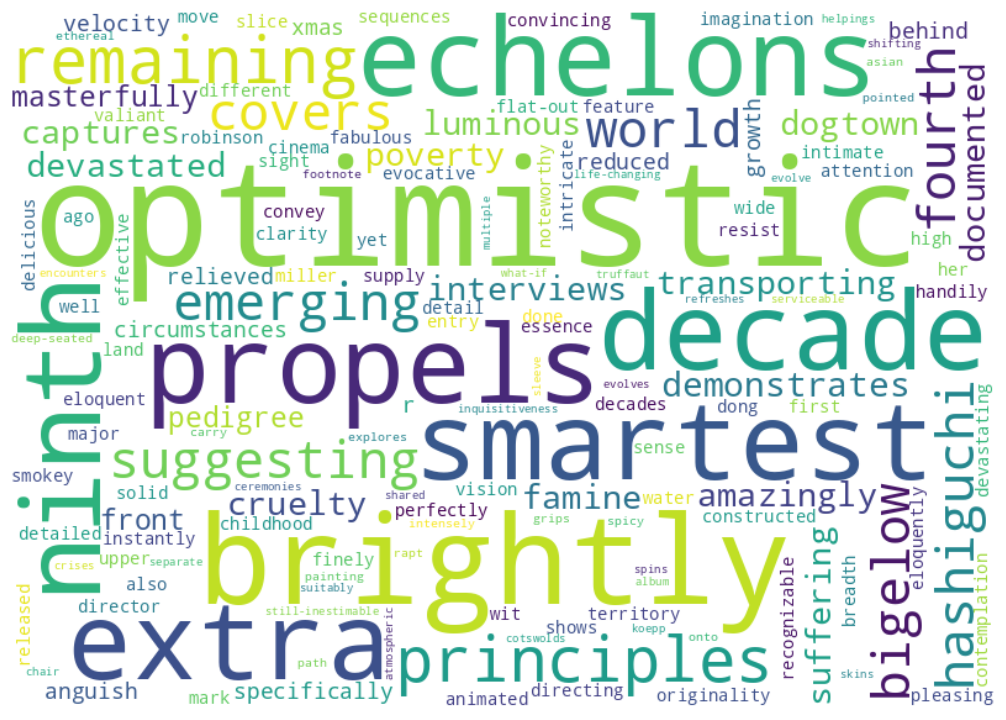}
\caption{ALBERT Layer 11 Cluster 489 (Positive Sentiment)}
\end{subfigure}
\begin{subfigure}[b]{0.30\linewidth}
\centering
\includegraphics[width=\linewidth]{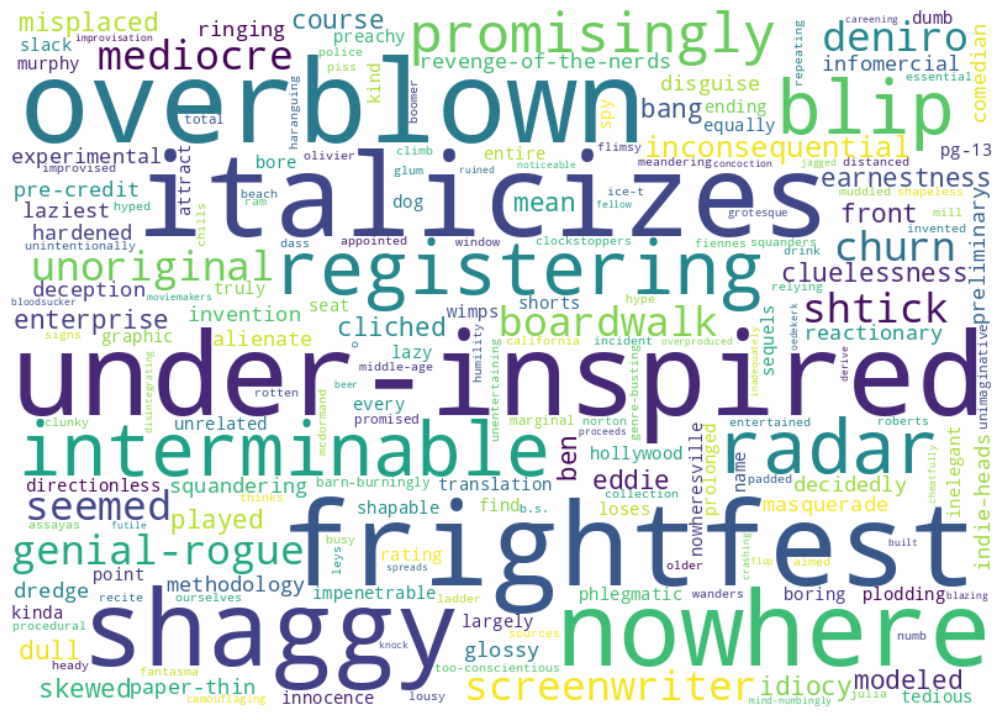}
\caption{ALBERT Layer 11 Cluster 215 (Negative Sentiment)}
\end{subfigure}
\caption{Task-specific latent clusters from various models and layers}
\label{fig:appendix-sample-clusters}
\end{figure*}

Figure \ref{fig:appendix-sample-clusters} shows some task-specific latent clusters from various models and layers.

\end{document}